\documentclass[journal, twoside]{IEEEtran/IEEEtran}

\usepackage{graphics} 
\usepackage{graphicx} 

\usepackage{epsfig} 
\usepackage{amsmath} 
\usepackage{amssymb}  

\usepackage[noadjust]{cite}

\usepackage{algorithm}
\usepackage{algpseudocode}

\usepackage[table]{xcolor}

\usepackage{booktabs}
\usepackage{multirow}
\usepackage[caption=false]{subfig}
\usepackage[draft]{hyperref}

\usepackage{fixltx2e}
\usepackage{url}

\usepackage{tikz}
\usetikzlibrary{shapes,snakes}
\usepackage{verbatim}
\usepackage{tkz-euclide}
\usetikzlibrary{arrows.meta}
\usetikzlibrary{calc}

\newtheorem{proposition}{\textbf{Proposition}}
\newtheorem{theorem}{\textbf{Theorem}}
\newtheorem{lemma}{\textbf{Lemma}}
\newtheorem{remark}{\textbf{Remark}}
\newtheorem{definition}{\textbf{Definition}}

\renewcommand{\IEEEQED}{\IEEEQEDopen}

\makeatletter

\newcommand{\Rmnum}[1]{\expandafter\@slowromancap\romannumeral #1@}
\makeatother

\DeclareMathOperator*{\argmin}{arg\,min}

	\usepackage{mathtools}
	
\makeatletter
\newcommand\fs@betterruled{%
	\def\@fs@cfont{\bfseries}\let\@fs@capt\floatc@ruled
	\def\@fs@pre{\vspace*{5pt}\hrule height.8pt depth0pt \kern2pt}%
	\def\@fs@post{\kern2pt\hrule\relax}%
	\def\@fs@mid{\kern2pt\hrule\kern2pt}%
	\let\@fs@iftopcapt\iftrue}
\floatstyle{betterruled}
\restylefloat{algorithm}
\makeatother

\begin{document}


\title{
	{Sparse Pose Graph Optimization in Cycle Space}
	}

\author{Fang Bai, \IEEEmembership{Member, IEEE},
	Teresa Vidal-Calleja, \IEEEmembership{Member, IEEE}, Giorgio Grisetti, \IEEEmembership{Member, IEEE}%
		\thanks{Fang Bai and Teresa Vidal-Calleja are with the Centre for Autonomous Systems (CAS), University of Technology Sydney, Sydney, Australia.
			Giorgio Grisetti is with the Department of Computer, Control, and Management Engineering ``Antonio Ruberti", Sapienza University of Rome, Rome, Italy.
			(E-mail:
			fang.bai@yahoo.com;
			Fang.Bai@student.uts.edu.au; Teresa.VidalCalleja@uts.edu.au; Grisetti@diag.uniroma1.it).
		}
}

\markboth{IEEE transactions on robotics}{FANG BAI \MakeLowercase{\textit{et al.}}:
	Sparse Pose Graph Optimization in Cycle Space.}
%

%


\maketitle


\begin{abstract}

The state-of-the-art modern pose-graph optimization (PGO) systems are vertex based.
In this context the number of variables might be high, albeit the number of cycles in the graph (loop closures) is relatively low.
For sparse problems particularly, the cycle space has a significantly smaller dimension than the number of vertices.
By exploiting this observation, in this paper we propose an alternative solution to PGO, that directly exploits the cycle space.
We characterize the topology of the graph as a cycle matrix, and re-parameterize the problem using relative poses, which are further constrained by a cycle basis of the graph.
We show that by using a minimum cycle basis, the cycle-based approach has superior convergence properties against its vertex-based counterpart, in terms of convergence speed and convergence to the global minimum.
For sparse graphs, our cycle-based approach is also more time efficient than the vertex-based.
As an additional contribution of this work we present an effective algorithm
to compute the minimum cycle basis.
Albeit known in computer science, we believe that this algorithm is not familiar to the robotics community.
All the claims are validated by experiments on both standard benchmarks and simulated datasets.
To foster the reproduction of the results, we provide a complete open-source C++ implementation\footnote{Code: \url{https://bitbucket.org/FangBai/cycleBasedPGO}}
of our approach.

\end{abstract}

\begin{IEEEkeywords}
	Pose graph optimization,
	minimum cycle basis,
	SE(3) manifold,
	simultaneous localization and mapping (SLAM)
\end{IEEEkeywords}

\section{Introduction}

\IEEEPARstart{P}{ose} graph optimization (PGO) is a fundamental problem which arises in various research disciplines, like
simultaneous localization and mapping (SLAM)
\cite{dissanayake2001solution, cadena2016past, mur2015orb, mur2017orb, schlegel2018proslam},
structure from motion
\cite{hartley2013rotation, arrigoni2016spectral, tron2016survey},
calibration of multi-camera rig \cite{esquivel2007calibration},
and sensor network localization
\cite{wang2010survey, peters2015sensor}.

A pose graph is a graph whose vertices encode positions and orientations of 3D poses,
and whose edges represent spatial constraints between the connected
vertices.
Taking a graph-based SLAM system as an example, the system processes the raw measurements to construct local maps.
These local maps are then arranged in a pose graph as vertices.
Constraints between local maps arise from matching nearby local
maps or from proprioceptive measurements coming from odometry or
inertial measurement units (IMU).

Typically the constraints are affected by some uncertainty,
which is modeled as a Gaussian (or Langevin) distribution centered around the
equilibrium point of the constraint.
For example, in graph-based SLAM,
systematic biases, noise in the sensors, errors in localization
propagate to the estimation of these constraints.
Hence, in real
applications it is impossible to find a configuration of vertices
that simultaneously nulls the residual error of all constraints.
PGO is then the task finding the configuration of the
vertices that is maximally consistent with the constraints
(i.e., edges), via solving a nonlinear least squares optimization problem.

In real SLAM applications, the number of edges is typically proportional to the
number of vertices. This is a consequence of the local nature of SLAM,
stemming from the limits in the sensor range. Only local maps that are
spatially close can share some common elements, and thus the
corresponding vertices can be connected by constraints (i.e., edges). This results
in limiting the number of edges connected to a vertex, and ultimately leads to a
sparsely connected graph.  By leveraging on this sparsity, modern
PGO systems \cite{dellaert2006square, kummerle2011g, ila2017slam++, rosen2019se} are capable of solving extremely large problems in a fraction of seconds.
We term the method in \cite{dellaert2006square, kummerle2011g, ila2017slam++, rosen2019se} as vertex-based approaches for a reason that will explain later on.

At its core, the state-of-the-art PGO techniques solve a sparse linear system to update the estimates of vertices \cite{dellaert2006square, kummerle2011g, ila2017slam++, rosen2019se}.
This system is typically solved by a sparse Cholesky factorization \cite{davis2006direct}, which is guided by the graph topology presented as a vertex-edge incidence matrix.
In graph theory, the incidence matrix spans a space called cut space, which is orthogonal complementary to a space called cycle space.
A sparse graph implies two facts: (a) low connectivity between vertices, which has been reflected in the incidence matrix and exploited effectively \cite{davis2006direct}; (b) low dimensionality of cycle space, which has been largely ignored due to the huge success of sparse Cholesky factorization with respect to vertices.
In this paper, we will show the possibility of designing effective PGO techniques in the cycle space.

To this end, we reformulate the conventional least squares optimization with a relative parametrization, i.e., using relative poses (associated with edges), as variables to be estimated.
This induces the over-parameterization of the problem since the number of edges is higher than that of vertices.
This issue is solved by introducing a collection of inherent constraints,
that the value of vertices anchors all the paths between any two vertices to generate the same composed transformation.
Such a set of constraints are topologically characterized by the graph cycle space,
which can be described by a cycle basis.
Finally, our optimization problem is casted as a constrained least squares optimization problem,
which can be solved in the sequential quadratic programming (SQP) scheme.
We term this PGO technique as cycle-based approach.

A common issue in the relative parameterization is that it is not necessarily sparse
\cite{huang2010far, bai2016incremental, bai2018robust, jackson2019direct}.
It turns out this issue can be resolved by using a minimum cycle basis (MCB); however the computation of a MCB itself is a hard problem.
We exploit the fact that graphs in PGO (and other similar applications) are sparse, with positive (integer) weights,
to design a tailored MCB algorithm that can greatly mitigate this issue, in particular for sparse graphs that are encountered in real SLAM/PGO applications.
It can be shown that relative formulations have faster convergence compared with the vertex-based ones in the absolute frame (both in this work and the work in \cite{jackson2019direct}).
Therefore for sparse graphs,
based on the MCB, the cycle-based approach can attain faster (or comparable at least) computational time compared with the vertex-based ones,
due to the reduced dimension in the cycle space and the sparsity forced by the MCB.
Aside from the numerical sparsity, the usage of the MCB can also improves the convergence to the global minimum.

Concretely, we make following contributions in this paper:
\begin{enumerate}
	\item We propose a cycle-based PGO formulation based on the cycle matrix and $\mathrm
	{SE}(3)$ Lie group, and derive a SQP algorithm on the manifold to solve it.
	\item We give insights that the matrix structure in the Cholesky factorization is characterized by a cycle matrix:
	(a) the matrix to be factorized has exactly the same dimension as that in the cycle space;
	(b) the numerical sparsity can be maximized by a MCB.
	\item We propose an effective MCB algorithm that is tailored for sparse graphs with positive integer weights.
	We give theoretical insights like LexDijkstra, and working heuristics like pruning vertices of degree two.
	\item We provide principled analyses in terms of observability, convergence with respect to cycle bases, and convergence rate for the cycle-based PGO formulation.
	\item We provide extensive experimental results to validate the advantages of using cycle-based PGO, in terms of both the computational complexity and robustness.
	\item We provide a C++ implementation of the overall algorithm which is freely available to the community.
		
\end{enumerate}

The remainder of this paper is structured as follows:
Section \ref{tro: section related work} reviews the related work.
The Lie group and graph preliminaries are provided in
Section \ref{tro: section: preliminaries and backgrounds}.
Section \ref{Pose-Graph as A Topological-metric Graph}
recaps the conventional vertex-based PGO formulation.
Section \ref{tro: section: cycle based pose-graph optimization}
derives the cycle-based PGO formulation and the corresponding SQP solver on the manifold.
Section \ref{tro: section: minimum length cycle basis}
is dedicated to calculate a MCB for sparse graphs with positive integer weights.
Analyses on the observability and convergence are presented in
Section \ref{section: analyses on cycle based PGO}.
Details of our C++ implementation are provided in
Section \ref{tro: section: implemenation details}.
Experimental validations are given in
Section \ref{tro: section: experimental results}.
Section \ref{tro: section: conlcusion}
concludes the paper.

\section{Related Work}
\label{tro: section related work}


PGO, as a maximum likelihood estimation (MLE), was firstly described in the seminal paper by Lu et al. \cite{lu1997globally},
where a nonlinear least squares (NLS) is used to optimize the network generated by scan-matchings.
At the time, although in theory, techniques like Gauss-Newton \cite{madsen1999methods} were available to solve the NLS problem,
the development of its numerical side was a bit behind.
To address the computational complexity,
Frese et al. \cite{frese2005multilevel} proposed a multi-level relaxation method, based on the Gauss-Seidel relaxation.
Olson et al. \cite{olson2006fast} suggested an incremental pose parameterization, and a PGO solver based on the stochastic gradient descent method, which had a large basin of convergence to the global optimum.
Grisetti et al. \cite{grisetti2009nonlinear} extended the framework to 3D, and applied
a tree parameterization to improve the convergence speed.


The rapid advancement in sparse linear algebra techniques (see Davis \cite{davis2006direct}) completely changed the landscape.
In robotics,
Dellaert et al. \cite{dellaert2006square} is the first to show that MLE can be solved efficiently by a Gauss-Newton method, using sparse matrix decompositions.
Kaess et al. attributed to the incremental solver of MLE,
using the Givens rotation based QR decomposition \cite{kaess2008isam},
or the Bayes tree \cite{kaess2012isam2}.
Kummerle et al. \cite{kummerle2011g} designed a general graph based optimization framework.
Ila et al. \cite{ila2017slam++} exploited the block structure of sparse matrices.
Besides matrix decompositions,
the resulting linear system can also be solved by an iterative method, for example preconditioned conjugate gradient \cite{konolige2004large, montemerlo2006large, dellaert2010subgraph}.
The convergence property of the Gauss-Newton based method can be further improved by using the idea of trust region \cite{madsen1999methods}, like Levenberg-Marquardt \cite{marquardt1963algorithm},
or Powell's-dog-leg \cite{rosen2014rise}.
All these MLE techniques can provide rather efficient PGO solutions.


It is possible to exploit specific structures of PGO to design more specialized solvers,
for example, the divide-and-conquer methods by
Grisetti et al. \cite{grisetti2010hierarchical, grisetti2012robust}.
The basic idea is to divide the full PGO into several submaps (i.e. subgraphs),
solve each one of them,
and then join the submaps together to obtain an approximation to the full PGO.
Zhao et al. \cite{zhao2013linear, zhao2019linear} investigated the special case of joining two submaps, with a clever parameterization which can be solved by a linear least squares,
followed by a nonlinear transformation.
For 2D cases,
Carlone et al. \cite{carlone2014fast} suggested a linear approximation framework to PGO,
by computing firstly an orientation estimation,
and then the position part using the given orientation.
The core was the regularization of rotation angles \cite{carlone2014fast},
which was systematically addressed in \cite{carlone2014angular} using a quadratic integer programming.
The separability of the orientation and position estimation was further studied by Khosoussi et al. \cite{khosoussi2016sparse},
based on a variable projection approach.


Besides practical algorithms to solve PGO, some theoretical insights are also drawn.
Huang et al. \cite{huang2010far} showed empirically that a point-feature based SLAM is close to a convex optimization problem, and a relative formulation is proposed for the purpose of reducing the nonlinearity in SLAM.
Wang et al. \cite{wang2013structure} discussed the number of local minimums for PGO in special cases.
Carlone \cite{carlone2013convergence} provided an analysis on the convergence basin of the global minimum for the Gauss-Newton method.
Several key factors are concluded, for example orientation noises, and graph topologies.
With the assumption of isometric additive Gaussian noise,
Khosoussi et al. \cite{khosoussi2018reliable} established the connection between the Fisher information matrix of the estimate and the graph complexity.
%
%


Convex optimization is a powerful technique to design globally optimal solutions to PGO.
An early touch on this topic came from Liu et al. \cite{liu2012convex},
who relaxed planar PGO into a semi-definite programming (SDP)
which was solved by standard convex optimization tools
\cite{boyd2004convex}.
It can be observed that the non-convexity of PGO comes from the cost function, and
manifold constraints.
Carlone and Rosen et al. showed that
the cost function can be chosen convex by using a Frobenius norm with an isotropic noise model
\cite{carlone2016planar, carlone2015duality, carlone2015lagrangian, rosen2015convex, rosen2019se}.
The manifold constraints can be relaxed by their convex-hulls, as shown by
Rosen et al. \cite{rosen2015convex}.
However, the relaxations in \cite{liu2012convex, rosen2015convex}
are not tight enough.
Carlone et al. \cite{carlone2016planar, carlone2015lagrangian}
explored the Lagrangian duality of PGO, leading to a tight SDP relaxation,
which can be verified to be globally optimal in many cases.
Rosen et al. \cite{rosen2019se, rosen2017computational} designed a certifiable PGO solver, exploiting convex relaxations, and a Riemannian trust-region method.
Briales et al. \cite{briales2017cartan}
suggested a compact matrix formulation, with concise and efficient derivations.


Loop-closing constraints and cycles have a rich history in SLAM literatures.
Estrada et al. \cite{estrada2005hierarchical} formulated the loop-closing problem between local maps as a quadratic optimization problem with equality constraints.
The problem was solved by SQP, and a connection to iterated extended Kalman filter was drawn.
Russell et al. \cite{russell2011optimal} proposed a distributed network optimization method based on the graph cycle space, and proved its convergence in linear cases.
%
%
%
Olson \cite{olson2008robust} evaluated the pairwise consistency of two loop-closing edges by joining them with the odometry sequence as a cycle.
Dubbelman et al. \cite{dubbelman2015cop} employed interpolations in $\mathrm{SE}(3)$ along a cycle to obtain an approximate solution to pose-chain SLAM.
The concept of cycle bases was used by Carlone et al. in \cite{carlone2014fast, carlone2014angular, carlone2013convergence}.	
The loop-closing cycle in a point-feature based SLAM was considered by Bai et al. \cite{bai2016incremental},
while both point and line-features are included by Guo et al. \cite{guo2016large} in more specific scenarios.
Later, Bai et al. \cite{bai2018robust} formulated PGO explicitly as a constrained optimization problem by using cycles in the graph.
The cycle structure in graph optimization is typically presented as relative formulations \cite{bai2016incremental, bai2018robust},
which have been used in the work
\cite{wang2005decoupling, olson2006fast, grisetti2007tree, grisetti2009nonlinear, sibley2009adaptive, huang2010far, anderson2015relative, moreno2016constant, jackson2019direct}
as well.

Although the usage of concepts like cycle space and cycle bases abounds in existing literatures,
none of these works study how to design efficient cycle-based optimization algorithms
by exploiting: the dimension reduction of Cholesky factorization in the cycle space due to graph sparsity, and the possibility of designing a tailored MCB algorithm that takes advantage of sparsity and positive integer weights.

	\section{Preliminaries and Backgrounds}
	\label{tro: section: preliminaries and backgrounds}

	\subsection{Notations}

	For any two sets $\mathcal{X}_1$ and $\mathcal{X}_2$, we denote respectively
	$\mathcal{X}_1 \cap \mathcal{X}_2$ the intersection,
	$\mathcal{X}_1 \cup \mathcal{X}_2$ the union,
	$\mathcal{X}_1 \backslash \mathcal{X}_2$ the difference,
	and
	$\mathcal{X}_1 \oplus \mathcal{X}_2 = (\mathcal{X}_1 \cup \mathcal{X}_2) \backslash (\mathcal{X}_1 \cap \mathcal{X}_2)$ the symmetric difference
	of these two sets.
	Let $\lvert \mathcal{X} \rvert$ be the cardinality of the set $\mathcal{X}$.
	We will use $\mathbb{Z}$ to denote the set of integers, and $\mathbb{R}$ the set of real numbers.
	If not explicitly stated, the lower-case in normal font, the lower-case in bold font, and the upper-case in bold font are reserved for scalars, vectors, and matrices respectively.
	A matrix of zeros is denoted by $\mathbf{O}$,
	and an identity matrix is denoted by $\mathbf{I}$.
	$\mathbf{A}^T$ represents the transpose,
	and $\mathbf{A}^{\dagger}$ the Moore-Penrose pseudo inverse of a matrix $\mathbf{A}$.
	The notation $\{\mathbf{v}_i\}^{\perp}$ stands for the orthogonal complement to the space spanned by a set of vectors $\{\mathbf{v}_i\}$.
	$<\mathbf{v}_1,\mathbf{v}_2> = \mathbf{v}_1^T \mathbf{v}_2$ is the inner product between $\mathbf{v}_1$ and $\mathbf{v}_2$.
	The squared Mahalanobis distance is denoted by
	$\lVert \mathbf{e} \rVert_{\boldsymbol{\Sigma}}^2 = \mathbf{e}^{T} \boldsymbol{\Sigma}^{-1} \mathbf{e}$.
	The notation $[m:n]$ is used to describe a sequence of consecutive integers from $m$ to $n$.
	We will use ``iff" as a shorthand of ``if and only if".
	The graph notations used throughout the paper are listed in Table \ref{table: list of noations on graph}.

\begin{table}[t]
	\caption{List of Notations on Graph}
	\label{table: list of noations on graph}
	\centering
	\begin{tabular}{lll}
	\toprule
		$ \mathbb{GF} = \mathbb{Z}_2  $   &           &   finite field of modulo $2$  \\
		$ \mathbb{GF}^{d_1 \times d_2}  $   &           &  $d_1 \times d_2$ dimensional matrix on $\mathbb{Z}_2$  \\		
	\midrule
		$\mathcal{G}  = \mathcal{G}(\mathcal{V}, \mathcal{E})$  &    & undirected graph    \\
		$\mathcal{E}$  &    & edge set of a graph $\mathcal{G}$  \\
		$\mathcal{V}$  &    & vertex set of a graph $\mathcal{G}$   \\	
		$e_{uv}$         &  $\mathcal{E}$   & edge with endpoints $u$, $v$   \\
		$ \nu = \lvert \mathcal{E} \rvert - \lvert \mathcal{V} \rvert + 1$  &  $\mathbb{R}$ &  dimension of cycle space  \\
	\midrule
		$\mathcal{T} = \mathcal{T}(\mathcal{G})$  &  $\mathbb{GF}^{1 \times m}$    & tree in $\mathcal{G}$   \\	
		$\mathcal{P}_{uv} = \mathcal{P}_{uv}(\mathcal{G})$  &  $\mathbb{GF}^{1 \times m}$   & path in $\mathcal{G}$, from vertex $u$ to $v$    \\
		$\mathcal{C} = \mathcal{C}(\mathcal{G})$  &  $\mathbb{GF}^{1 \times m}$   & cycle in $\mathcal{G}$   \\
		$\mathcal{S} $ &   $\mathbb{GF}^{1 \times m}$   & support vector   \\  
		\multicolumn{3}{l}{Note: A vector on $\mathbb{GF}^{1 \times m}$ can be sparsely described by a set.} \\	
	\midrule
		$\{ * \}$ &   &  set with elements in form of $*$ \\
		$\boldsymbol{\mathcal{H}}$	&  $\{\mathcal{C}\}$  &  Horton set  \\
		$\boldsymbol{\mathcal{I}}$	&  $\{\mathcal{C}\}$  &  set of isometric cycles  \\
		$\boldsymbol{\mathcal{B}}$	&  $\{\mathcal{C}\}$  &  cycle basis  \\
		$\mathcal{B} = \mathcal{B}(\mathcal{G})$  &  $\mathbb{GF}^{n \times \nu}$   & cycle matrix of cycle basis $\boldsymbol{\mathcal{B}}$    \\
	\midrule
		$\mathbf{T}_k$     &  $\mathrm{SE(3)}$  &   poses \\
		$\mathbf{T}_{\mathbf{k}}$     &  $\mathrm{SE}(3)$ &   relative poses  \\
		$\boldsymbol{\mathcal{P}}_{uv}$ &  $\{\mathbf{T}_{\mathbf{k}}\}$    & geometric path from $u$ to $v$  \\
		$\boldsymbol{\mathcal{C}}$   &  $\{\mathbf{T}_{\mathbf{k}}\}$    & geometric cycle   \\
	\bottomrule
	\end{tabular}
\end{table}

	\subsection{Special Euclidean Group}

	The special Euclidean group, $\mathrm{SE}(3)$, is a standard tool to describe rigid body transformations \cite{chirikjian2011stochastic, barfoot2017state},
	which typically occur in robotics and computer vision community.

	$\mathrm{SE}(3)$ is a Lie group.
	A Lie group is a peculiar smooth manifold whose local structure can be described by
	the so-called Lie algebra, which is the tangent space at the identity of the group.
	Let $\mathfrak{se}(3)$ be the Lie algebra of $\mathrm{SE}(3)$.
	Both $\mathrm{SE}(3)$ and $\mathfrak{se}(3)$ can be described by matrices.
	For each matrix $\mathbf{T} \in \mathrm{SE}(3)$, we can find an associated matrix $\mathbf{X} \in \mathfrak{se}(3)$, and vice versa, by matrix exponential and matrix logarithm:
	$\mathbf{T} = \mathbf{exp} ( \mathbf{X} )$,
	$\mathbf{X} = \mathbf{log} ( \mathbf{T} )$.

	An element $\mathbf{X} \in \mathfrak{se}(3)$ can be identified by a ``screw matrix", taking the form
	\begin{equation*}
	\mathbf{X}
	=
	\begin{bmatrix}
	0 & -x_3 & x_2 & x_4 \\
	x_3 & 0 & -x_1 & x_5 \\
	-x_2 & x_1 & 0 & x_6 \\
	0 & 0 & 0 & 0
	\end{bmatrix}
	.
	\end{equation*}
	It is obvious that each screw matrix $\mathbf{X}$ can be uniquely identified by a vector
	$
	\mathbf{x} = 
	[x_1, x_2, x_3, x_4, x_5, x_6]^{T}
	\in \mathbb{R}^6
	$.
	The relationship can be expressed as
	$
	\mathbf{X} =\mathbf{x}^{\wedge}
	$,
	$
	\mathbf{x} =\mathbf{X}^{\vee}
	$
	with operation $\wedge$ and $\vee$.
	Therefore for convenience, we define an encapsulated exponential and logarithm mapping between $\mathbf{T}$ and $\mathbf{x}$ directly as
	\begin{equation*}
		\mathbf{T} = \mathbf{Exp} (\mathbf{x}) = \mathbf{exp} (\mathbf{x}^{\wedge})
		,\ 
		\mathbf{x} = \mathbf{Log} (\mathbf{T}) = \mathbf{log} (\mathbf{T}^{\vee})
		.
	\end{equation*}
	
	The adjoint of Lie algebra, $\mathbf{ad}(\mathbf{x})$,
	is related to a binary operation $[\cdot,\cdot]$, called Lie bracket,
	yielding the relation
	\begin{equation*}
	[\mathbf{X},\mathbf{Y}] 
	= \mathbf{X} \mathbf{Y} - \mathbf{Y} \mathbf{X}
	= ( \mathbf{ad}(\mathbf{x}) \mathbf{y} )^{\wedge}
	,
	\end{equation*}
	which holds for any $\mathbf{Y} \in \mathfrak{se}(3)$,
	$\mathbf{y} = \mathbf{Y}^{\vee} \in \mathbb{R}^6$.
	Exponentiating $\mathbf{ad}(\mathbf{x})$, we would get a matrix $\mathbf{Ad} ( \mathbf{T} )$ called the adjoint of Lie group.
	The adjoint matrix has a nice property,
	\begin{equation}
	\label{Adjoint Property of Lie Group}
	\mathbf{T} \cdot \mathbf{Exp} (\mathbf{y}) 
	=
	\mathbf{Exp} \left( \mathbf{Ad} \left( \mathbf{T} \right) \mathbf{y} \right)
	\cdot \mathbf{T}
	\end{equation}
	which can be used to shift the position of $\mathbf{T}$ and $\mathbf{Exp}(\cdot)$.
	Another property of $\mathbf{Ad}(\cdot)$ is
	\begin{equation*}
	\mathbf{Ad} (\mathbf{T}_1)
	\mathbf{Ad} (\mathbf{T}_1)
	=
	\mathbf{Ad} (\mathbf{T}_1 \mathbf{T}_2)
	\end{equation*}
	which is used to collect two $\mathbf{Ad}(\cdot)$ together.

	 The Baker-Campbel-Hausdorff formula (BCH) is used to concatenate two matrix exponentials.
	The exact BCH formula is expressed as a series, and a closed form approximation is:
	\begin{equation*}
	\mathbf{Exp}\left(\mathbf{x}\right) \cdot \mathbf{Exp}\left(\mathbf{y}\right)
	\approx
	\begin{cases}
	\mathbf{Exp}\left(\mathbf{J}_{\mathbf{l}}^{-1}\left(\mathbf{y}\right) \mathbf{x} + \mathbf{y} \right)
	,\ \ 
	\mathrm{if}\ \mathbf{x}\rightarrow\mathbf{0}
	\\
	\mathbf{Exp}\left(\mathbf{x} +  \mathbf{J}_{\mathbf{r}}^{-1}\left(\mathbf{x}\right) \mathbf{y} \right)
	,\ \ 
	\mathrm{if}\ \mathbf{y}\rightarrow\mathbf{0}
	\end{cases}
	\end{equation*}
	where $\mathbf{J}_{\mathbf{l}}(\cdot)$, $\mathbf{J}_{\mathbf{r}}(\cdot)$ are called the left-hand and right-hand Jacobian of the exponential coordinate parameterization.

	The mappings between $\mathrm{SE}(3)$ and $\mathfrak{se}(3)$,
	the adjoint operation,
	and the BCH formula are used to linearize PGO, which is the prerequisite to apply an iterative nonlinear solver.

	For $\mathrm{SE(3)}$, all the operations
	$\mathbf{Exp}(\cdot)$, $\mathbf{Log}(\cdot)$,
	$\mathbf{ad}(\cdot)$, $\mathbf{Ad}(\cdot)$,
	$\mathbf{J}_{\mathbf{l}}(\cdot)$ and $\mathbf{J}_{\mathbf{r}}(\cdot)$
	are calculated in closed form \cite{chirikjian2011stochastic, barfoot2017state}.

	\subsection{Graph Theory}

	Let $\mathcal{G}$ be an undirected graph ${\mathcal{G}}(\mathcal{V}, \mathcal{E})$,
	where $\mathcal{V}$ is a finite set,
	and $\mathcal{E}$ is a set of unordered pairs $(u,v)$, with $u,v \in \mathcal{V}$.
	The elements in $\mathcal{V}$ are termed \textsf{vertices} (or nodes), and the elements in $\mathcal{E}$ are termed \textsf{edges}.
	In what follows, we will denote an edge from $u$ to $v$ as $e_{uv}$.
	An edge $e_{uv}$ is said to be \textsf{incident} to the vertices $u$ and $v$,
	while $u$ and $v$ are called the \textsf{endpoints} of $e_{uv}$.
	The \textsf{degree} of a vertex in $\mathcal{G}$ is the number of edges incident to that vertex.
	A \textsf{subgraph} of $\mathcal{G}$ stands for a graph with only part of vertices and edges from $\mathcal{G}$.
	In particular, we will be interested in three types of \textsf{subgraphs}, i.e., \textsf{path}, \textsf{cycle}, and \textsf{tree}.
	Formally, a graph is said to be \textsf{connected} if there exists a path for any pair of vertices in the graph.
	A \textsf{path} is a connected subgraph in which there are exactly two vertices having degree of one, and the rest of vertices having degree of two.
	A \textsf{cycle} is a subgraph in which every vertex has an even degree.
	If a cycle is connected and the degree of each vertex is exactly two, the cycle is called a \textsf{simple/elementary cycle}, or a \textsf{circuit}.
	A \textsf{tree} is a connected subgraph which contains no cycles (i.e. \textsf{acyclic} subgraph).
	If a tree of $\mathcal{G}$ contains all vertices in $\mathcal{V}$, it is called a \textsf{spanning tree} of $\mathcal{G}$.
	We will use
	$\mathcal{P}$ to denote a path,
	$\mathcal{C}$ a cycle,
	and $\mathcal{T}$ a tree,
	respectively.

	A subgraph, i.e., a path/cycle/tree, can be uniquely identified by the set of edges it used,
	which induces an ``incidence vector" whose elements are assigned to either $0$ or $1$.
	For instance, a cycle $\mathcal{C}$ can be expressed as an incidence vector
	$[c_1, c_2, \dots, c_{\lvert \mathcal{E} \rvert} ]$,
	with $c_{k} = 1$ ($k = 1, 2, \dots \lvert \mathcal{E} \rvert$) iff the $k$-th edge is used by the cycle $\mathcal{C}$, and $c_{k} = 0$ otherwise (see Fig. \ref{fig An illustration of incidence matrix and cycle matrix of a graph}).
	The concept of \textsf{finite field} (or \textsf{Galois field}) is useful to describe this phenomenon.
	A finite field is basically a finite set equipped with arithmetic rules.
	In particular, we are interested in the finite field of order $2$,
	denoted as $\mathbb{GF} = \mathbb{Z}_2 = \{0, 1\}$, whose elements are $0$ and $1$ only.
	The addition and multiplication on $\mathbb{Z}_2$ are defined respectively to be the addition and multiplication on $\mathbb{Z}$ modulo $2$.
	Obviously, incidence vectors (to describe paths/cycles/trees) are vectors on $\mathbb{GF}$.
	Moreover, all the cycles in $\mathcal{G}$ can be described by a matrix on $\mathbb{GF}$ with each row being a cycle incidence vector.
	This matrix is called \textsf{cycle matrix}: $\mathcal{B} = [b_{i,j}]$,
	with $b_{i,j} = 1$ iff the $j$-th edge is contained in the $i$-th cycle, and $b_{i,j} = 0$ otherwise.
	The rows of $\mathcal{B}$ span a vector space over the two-element \textsf{finite field} based on the modulo two arithmetics, which is called \textsf{cycle space}.
	A basis to the cycle space is called \textsf{cycle basis}.
	The cycle space of an undirected graph is orthogonal complementary to the so-called \textsf{cut space}.
	The cut space is not needed to understand this work, but important to build connections with Gauss-Newton based optimizers \cite{dellaert2006square, kummerle2011g, ila2017slam++}. Interested readers are referred to \cite{ray2012graph, kavitha2009cycle} for accessible explanations.
	For a connected graph, the cycle space has a dimension $\nu = \lvert \mathcal{E} \rvert - \lvert \mathcal{V} \rvert + 1$, and the cut space has a dimension $\lvert \mathcal{V} \rvert - 1$.

	Let $\mathbf{x}_1$, $\mathbf{x}_2$ be two vectors on $\mathbb{GF}^{\lvert \mathcal{E} \rvert}$,
	and $\mathcal{X}_1$, $\mathcal{X}_2$ be the corresponding set representations.
	Then the vector addition $\mathbf{x}_1 + \mathbf{x}_2$ on $\mathbb{GF}^{\lvert \mathcal{E} \rvert}$ corresponds to the symmetric difference of sets, i.e., $\mathcal{X}_1 \oplus \mathcal{X}_2$.
	The inner product satisfies:
	$< \mathbf{x}_1, \mathbf{x}_2 > = \mathbf{x}_1^T \cdot \mathbf{x}_2 = 1$
	iff $\lvert \mathcal{X}_1 \cap \mathcal{X}_2 \rvert$ is odd;
	$< \mathbf{x}_1, \mathbf{x}_2  > = \mathbf{x}_1^T \cdot \mathbf{x}_2 = 0$
	iff $\lvert \mathcal{X}_1 \cap \mathcal{X}_2 \rvert$ is even.
	In what follows,
	we will use the vector representation and set representation interchangeably
	and describe both with the same notation, where the difference can be easily told by the operation used.

	\begin{figure}[t]
		\centering
		\begin{tikzpicture}
		
		\def \xgridlen{2.0}
		\def \rotangle{75} 
		\def \ygridlen{\xgridlen * sin{\rotangle}}
		\def \dx{\xgridlen * cos{\rotangle}}

		\tikzset{VertexStyle1/.style = {
				shape = circle,
				draw=black!80, 
				fill= cyan!60,
				thick,
				solid,
				inner sep = 2pt,
				outer sep = 2pt,				
				minimum size   = 20 pt}}
		
		\tikzset{EdgeStyle1/.style   = {
				line width=1.2pt }}
		\tikzset{EdgeStyle2/.style   = {
				line width=1.2pt,
				->,
				> = stealth }}

		\tikzset{NodeStyle1/.style = {
				thick,
				solid,
				minimum size   = 5 pt}}

		\node[VertexStyle1] at (\xgridlen, 0) (v1) {$v_1$};
		\node[VertexStyle1] at (2*\xgridlen, 0) (v2) {$v_2$};
		\node[VertexStyle1] at (3*\xgridlen, 0) (v3) {$v_3$};
		\node[VertexStyle1] at (3*\xgridlen + \dx, \ygridlen) (v4) {$v_4$};
		\node[VertexStyle1] at (2*\xgridlen + \dx, \ygridlen) (v5) {$v_5$};
		\node[VertexStyle1] at (1*\xgridlen + \dx, \ygridlen) (v6) {$v_6$};
		\node[VertexStyle1] at (0*\xgridlen + \dx, \ygridlen) (v7) {$v_7$};

		\draw[EdgeStyle2] (v1) -- (v2) node[NodeStyle1, midway, above]{$e_{12}$}
		node[NodeStyle1, midway, below]{$\mathbf{T}_{12}$};
		\draw[EdgeStyle2] (v2) -- (v3) node[NodeStyle1, midway, above]{$e_{23}$}
		node[NodeStyle1, midway, below]{$\mathbf{T}_{23}$};
		\draw[EdgeStyle2] (v4) -- (v5) node[NodeStyle1, midway, above]{$e_{45}$}
		node[NodeStyle1, midway, below]{$\mathbf{T}_{45}$};
		\draw[EdgeStyle2] (v5) -- (v6) node[NodeStyle1, midway, above]{$e_{56}$}
		node[NodeStyle1, midway, below]{$\mathbf{T}_{56}$};
		\draw[EdgeStyle2] (v6) -- (v7) node[NodeStyle1, midway, above]{$e_{67}$}
		node[NodeStyle1, midway, below]{$\mathbf{T}_{67}$};
		\draw[EdgeStyle2] (v2) -- (v5) node[NodeStyle1, midway, above, sloped]{$e_{25}$}
		node[NodeStyle1, midway, below, sloped]{$\mathbf{T}_{25}$};
		\draw[EdgeStyle2] (v1) -- (v6) node[NodeStyle1, midway, above, sloped]{$e_{16}$}
		node[NodeStyle1, midway, below, sloped]{$\mathbf{T}_{16}$};
		\draw[EdgeStyle2] (v3) -- (v4) node[NodeStyle1, midway, above, sloped]{$e_{34}$};
		\draw[EdgeStyle2] (v3) -- (v4) node[NodeStyle1, midway, below, sloped]{$\mathbf{T}_{34}$};
		\end{tikzpicture}

		\begin{equation*}
		\mathcal{B}  = 
		\begin{array}{c|cccccccc}
		& e_{12} & e_{23} & e_{34} & e_{45} & e_{56} & e_{67} & e_{25} & e_{16} \\
		\midrule
		\mathcal{C}_1   & 1 &   &   &   & 1 &   & 1 & 1 \\
		\mathcal{C}_2   &   & 1 & 1 & 1 &   &   & 1 &   \\		
		\end{array}	
		\end{equation*}	
		\caption{A toy graph of PGO.
			For each relative poses $\mathbf{T}_{i,j}$, the edge $e_{ij}$ is oriented as $i \rightarrowtail j$.
			When talking about topological information, like cycles/paths, we can safely operate on the undirected version by ignoring the edge orientations, and lifting back to oriented edges when the cycles/paths are computed.
			In a graph, paths/cycles are a collection of edges, which can be described by a set or a vector on $\mathbb{GF}$.
			For example, there are three simple cycles in this graph:
			$\mathcal{C}_1 =\{ e_{12}, e_{25}, e_{56}, e_{16} \}$,
			$\mathcal{C}_2 =\{ e_{23}, e_{34}, e_{45}, e_{25} \}$,
			and
			$\mathcal{C}_3 =\{ e_{12}, e_{23}, 			
			e_{34}, e_{45}, e_{56}, e_{16} \}$.
			Cycles can be concatenated by the symmetric difference of sets:
			$\mathcal{C}_3 = \mathcal{C}_1 \oplus \mathcal{C}_2$.
			In this graph, $ \mathcal{C}_1$ and $\mathcal{C}_2$ are two independent cycles forming a cycle basis of the graph.
			The vectorized representation on $\mathbb{GF}$, i.e., the cycle basis matrix,
			is presented in $\mathcal{B}$, where the blanks are zeros.
			$\mathcal{C}_3$ can be written as a vector
			$\mathcal{C}_3  = [0, 1, 1, 1, 1, 1, 0, 0, 1]$.
			Based on the arithmetics of $\mathbb{GF}$, we have
			$\mathcal{C}_3 = \mathcal{C}_1 + \mathcal{C}_2$, which is in accordance with the symmetric difference of sets.
		}
		\label{fig An illustration of incidence matrix and cycle matrix of a graph}
	\end{figure}

	\section{Traditional Pose-Graph Optimization}
	\label{Pose-Graph as A Topological-metric Graph}

	The topology of PGO can be visualized as a graph whose vertices represent \textsf{poses}.
	An edge is created if there exists a relative geometric relation between two poses, i.e. \textsf{relative poses}, which can be produced by a wheel-encoder,
	scan-matching \cite{lu1997globally, besl1992method, segal2009generalized}, epipolar geometry \cite{hartley2003multiple, scaramuzza2011visual},
	or visual loop-closing techniques etc. \cite{cummins2008fab, lowry2016visual}.

	Both poses and relative poses are rigid-body transformations, which can be described by $\mathrm{SE(3)}$ Lie group.
	Denote $\mathbf{T}_i$ to be the $i$-th pose.
	The relative poses from the $i$-th pose to the $j$-th pose, denoted by $\mathbf{T}_{i,j}$, is a rigid-body transformation evaluated in the local coordinate frame of the $i$-th pose,
	which mathematically writes
	$\mathbf{T}_{i,j} = \mathbf{T}_i^{-1}  \mathbf{T}_j$.
	For the clarity of notations, we assign each edge a unique index, and use $\mathbf{T}_{\mathbf{k}}$ with subscript in bold to represent a relative poses.

	For each relative pose $\mathbf{T}_{\mathbf{k}}$,
	there is a noisy measurement $\mathbf{\tilde{T}}_{\mathbf{k}}$.	
	The measurement noise is conventionally assumed to be zero-mean Gaussian in the vector space of $\mathrm{SE(3)}$ Lie algebra, which can be mathematically formalized as,
	\begin{equation*}
	\mathbf{Log} (
	\mathbf{\tilde{T}}_{\mathbf{k}}^{-1} \cdot \mathbf{T}_{\mathbf{k}}
	)
	\sim \mathcal{N}(\mathbf{0}, \boldsymbol{\Sigma}_{\mathbf{k}})
	.
	\end{equation*}
	Note that other noise models are also possible, for example, the matrix Langevin distributions in \cite{carlone2016planar, rosen2019se}.

	Let the topology of PGO be described by the graph $\mathcal{G}(\mathcal{V}, \mathcal{E})$.
	Then PGO aims to obtain a maximum likelihood (MLE) estimator for the set of poses
	$\{ \mathbf{T}_i \}_{ i \in \mathcal{V} }$
	using the set of \textsf{measurements of relative poses}
	$\{ \mathbf{\tilde{T}}_{\mathbf{k}} \}_{ \mathbf{k} \in \mathcal{E} }$,
	via solving a least squares optimization problem,
	\begin{equation}
	\label{NLLS_PGO}
	\{ \mathbf{T}_i \}_{ i \in \mathcal{V} } 
	=
	\argmin \sum_{ \mathbf{k} \in \mathcal{E} } 
	\lVert   
	\mathbf{Log}
	(
	\mathbf{\tilde{T}}_{\mathbf{k}}^{-1} \cdot \mathbf{T}_{\mathbf{k}}
	)
	\rVert_{\boldsymbol{\Sigma}_{\mathbf{k}}}^2
	.
	\end{equation}

	\begin{remark}
		Note that the relative poses $\mathbf{T}_{i,j}$ is evaluated at the local frame of the $i$-th pose, so ideally a PGO is described by a directed graph.
		However, the edge orientation will not affect the topological side of a graph, like paths/cycles we discuss later on.
		Therefore we opt to describe PGO as an undirected graph $\mathcal{G}(\mathcal{V}, \mathcal{E})$.
		The restriction to the undirected graph limits the graph matrices/vectors to Galois field instead of real numbers.
		Besides, the undirected edges can be easily lifted to directed edges whenever it is desired.
			
	\end{remark}

	\begin{remark}
		\label{remark traditional pgo is unobservable}
		The traditional PGO is unobservable,
		in the sense that if $\{ \mathbf{T}_i \}_{ i \in \mathcal{V} }$ is a solution to (\ref{NLLS_PGO}), then $\forall \mathbf{T}' \in \mathrm{SE(3)}$,
		$\{ \mathbf{T}' \mathbf{T}_i \}_{ i \in \mathcal{V} }$ is also a solution to (\ref{NLLS_PGO}).
		This can be easily verified by the fact that
		$\mathbf{T}_{i,j} = \mathbf{T}_i^{-1}  \mathbf{T}_j = (\mathbf{T}' \mathbf{T}_i)^{-1}  (\mathbf{T}' \mathbf{T}_j)$,
		thus $\{ \mathbf{T}_i \}_{ i \in \mathcal{V} }$ and $\{ \mathbf{T}' \mathbf{T}_i \}_{ i \in \mathcal{V} }$ yields exactly the same contribution in the cost function.
		To ensure a unique solution, a popular practice is to anchor some poses (usually the first pose) to a fixed value or the identity of $\mathrm{SE(3)}$.

	\end{remark}

	\section{Cycle Based Pose-Graph Optimization}
	\label{tro: section: cycle based pose-graph optimization}

%
%

		The state-of-the-art PGO techniques \cite{dellaert2006square, kummerle2011g, ila2017slam++, carlone2016planar, rosen2019se}
		describe PGO as a factor graph \cite{dellaert2017factor},
		whose topology is represented by an incidence matrix.
		The graph is solved by a second-order optimization technique, for example Gauss-Newton, which results to solve a sparse linear system whose dimension is decided by the number of vertices.
		These approaches solve PGO in the cut space, and we will term them as vertex-based PGO (VB-PGO).
		The VB-PGO can be solved rather efficiently by sparse matrix factorizations \cite{davis2006direct}.

		Practical PGO instances are rather sparse.
		Let us measure the graph sparsity as a concept called \textsf{cycle ratio}, defined as $\frac{\nu}{\lvert \mathcal{E} \rvert}$, i.e., the dimension of the cycle space divided by the number of edges.
		Empirically, a PGO instance encountered in SLAM has a cycle ratio well below $20\%$,
		which implies $\frac{\nu}{\lvert \mathcal{V} \rvert -1} < 1/4$, i.e., the dimension of the cut space is at least four times larger than that of the cycle space.
		Let alone SLAM instances with a long trajectory and a few loop-closures,
		whose cycle ratio can be less than $5\%$, or even $1\%$.

		In this section, we will describe an approach that transforms PGO from the cut space to the cycle space.
		The cycle-based PGO, denoted as CB-PGO, has a reduced dimension compared with its vertex-based counterpart, as long as the graph is sparse enough.
		While the dimension reduction to the cycle space can undermine the sparsity of PGO, we propose to maximize the sparsity by a minimum cycle basis which will be described in Section \ref{tro: section: minimum length cycle basis}.
		The observability and convergence properties of the proposed CB-PGO are discussed in Section \ref{section: analyses on cycle based PGO}.

	\subsection{Preliminaries}

	In what follows, we will use the term
	\textsf{toplogical path} and \textsf{toplogical cycle}
	to represent a path and cycle in a pure topological graph $\mathcal{G}$.
	If such a $\mathcal{G}$ is associated with geometric information, namely by associating vertices with poses and edges with relative poses respectively, we will term this graph as a \textsf{geometric graph}.
	A (topological) path $\mathcal{P}$ whose edges are associated with relative poses will be termed
	as a \textsf{geometric path}, denoted by $\boldsymbol{\mathcal{P}}$.
	Analogously, a (topological) cycle $\mathcal{C}$ with edges associated with relative poses will be termed as a \textsf{geometric cycle}, denoted by $\boldsymbol{\mathcal{C}}$.
	If not explicitly stated, the terms, paths and cycles, refer to the topological version.

	The orientations of edges are irrelevant in this paper when discussing the topology of $\mathcal{G}$, as well as concepts like cycle bases and sparsity.
	However, they are useful in terms of describing the geometric paths/cycles.
	The \textsf{orientation of an edge} is decided by the geometric information, i.e., relative poses it associated with.
	For example, given an edge $\mathbf{k}$ in $\mathcal{G}$ with the associated relative poses being
	$\mathbf{T}_{i,j} = \mathbf{T}_i^{-1}  \mathbf{T}_j$,
	we stipulate the edge orientation to be from $i$ to $j$.
	In other words,
	$i \rightarrowtail j$ is the forward direction of the edge $\mathbf{k}$,
	and $j \rightarrowtail i$ is the backward direction.

	\subsection{Consistency of Pose-Graph Optimization}

	Let $\mathcal{P}_{st}$ be a path from $s$ to $t$ in $\mathcal{G}$.
	The corresponding \textsf{geometric path} $\boldsymbol{\mathcal{P}}_{st}$ is defined as: 
	\begin{equation}
	\label{equation: geometric path from s-t}
		\boldsymbol{\mathcal{P}}_{st}
		=
		\mathbf{T}_{\mathbf{1}^p}^{\varrho(\mathbf{1}^p)} 
		\mathbf{T}_{\mathbf{2}^p}^{\varrho(\mathbf{2}^p)} 
		\cdots 
		\mathbf{T}_{\boldsymbol{\theta}^p}^{\varrho(\boldsymbol{\theta}^p)}	
	\end{equation}
	where $\mathbf{T}_{\mathbf{1}^p}, \mathbf{T}_{\mathbf{2}^p}, \cdots, \mathbf{T}_{\boldsymbol{\theta}^p}$ is the sequence of relative poses by traversing $\mathcal{P}_{st}$ from $s$ to $t$.
	In (\ref{equation: geometric path from s-t}),
	the superscript	$\varrho(\mathbf{k}^p)$ is assigned to $+1$ if the traversal uses the edge $\mathbf{k}^p$ in the forward direction, and $-1$ if in the backward direction.
	The superscript of a matrix $\mathbf{T}$ will be interpreted as the power of the matrix, by
	$\mathbf{T}^{+1} = \mathbf{T}$
	and $\mathbf{T}^{-1} = \mathbf{inv} (  \mathbf{T} )$.

	PGO is a consistent formulation with respect to poses and relative poses.
	To show this, let $\mathbf{T}_s$ and $\mathbf{T}_t$ be two arbitrary poses.
	Let $\boldsymbol{\mathcal{P}}_{1:st}$ and $\boldsymbol{\mathcal{P}}_{2:st}$ be two geometric paths from pose $s$ to $t$.
	Then the pose $\mathbf{T}_t$ calculated from these two paths are exactly the same, i.e.,
	$\mathbf{T}_t = \mathbf{T}_s \boldsymbol{\mathcal{P}}_{1:st} = \mathbf{T}_s \boldsymbol{\mathcal{P}}_{2:st}$.
	Obviously, the consistency of the PGO can be also interpreted as the equivalence of the geometric paths, in the sense that
	$\boldsymbol{\mathcal{P}}_{1:st} =  \boldsymbol{\mathcal{P}}_{2:st} = \mathbf{T}_s^{-1} \mathbf{T}_t$.
	At last, with a slightly abuse of notation, we can write the consistency of two geometric paths as 
	$\boldsymbol{\mathcal{P}}_{1:st}^{-1}  \boldsymbol{\mathcal{P}}_{2:st} = \mathbf{I}$,
	which is a \textsf{geometric cycle}.
	
	The geometric cycles will play a key role in formulating PGO in cycle space, which in general ensures the consistency of the PGO.
	On the other hand, the underlying topological cycles will decide the sparsity of the proposed PGO formulation.

	\subsection{PGO in Cycle Space}

		Alternatively, we can traverse edges sequentially along a topological cycle, and consider the associated relative poses, to obtain a geometric cycle.
		For example, consider a topological cycle with length $\boldsymbol{\lambda}$.
		Let the sequential relative poses along the cycle be $\mathbf{T}_{\mathbf{1}^c}, \mathbf{T}_{\mathbf{2}^c}, \cdots, \mathbf{T}_{\boldsymbol{\lambda}^c}$,
		and the corresponding orientations of the edges be
		$ \sigma(\mathbf{1}^c), \sigma(\mathbf{2}^c), \cdots, \sigma(\boldsymbol{\lambda}^c) $,
		where $\sigma(\mathbf{k}^c)$ takes value $+1$ if the traversal uses the edge $\mathbf{k}^c$ in the forward direction, and $-1$ otherwise.
		Then the corresponding geometric cycle can be written as follow,
		\begin{equation*}
		{\boldsymbol{\mathcal{C}}}^{\mathrm{lhs}}
		= \mathbf{I},
		\quad \mathrm{with}\ \ 
		{\boldsymbol{\mathcal{C}}}^{\mathrm{lhs}}
		=
		\mathbf{T}_{\mathbf{1}^c}^{\sigma(\mathbf{1}^c)} 
		\mathbf{T}_{\mathbf{2}^c}^{\sigma(\mathbf{2}^c)} 
		\cdots 
		\mathbf{T}_{\boldsymbol{\lambda}^c}^{\sigma(\boldsymbol{\lambda}^c)}
		\end{equation*}
	where 
	$\mathbf{T}^{+1} = \mathbf{T}$
	and
	$\mathbf{T}^{-1} = \mathbf{inv} (  \mathbf{T} )$
	respectively.

	Based on the edge orientations and associated relative poses,
	for each topological cycle $\mathcal{C}$, we can generate a corresponding geometric cycle ${\boldsymbol{\mathcal{C}}}^{\mathrm{lhs}} = \mathbf{I}$.
	To characterize the cycle space of the graph $\mathcal{G}$, we need $\nu$ independent topological cycles, i.e., a cycle basis of $\mathcal{G}$.
	Let such a cycle basis be $\boldsymbol{\mathcal{B}} =\{ \mathcal{C}_i \}_{i = [1:\nu]}$, and its corresponding cycle matrix be $\mathcal{B}$.
	Then given the cycle basis $\boldsymbol{\mathcal{B}}$, we can find $\nu$ independent geometric cycles accordingly, denoted as
	$
	\{  {\boldsymbol{\mathcal{C}}}_i^{\mathrm{lhs}} = \mathbf{I}  
	 \}_{ i = [1:\nu] }
	$.

	Then let us
	take all $\vert \mathcal{E} \vert$ relative poses as new variables to be estimated,
	and take $\nu$ independent geometric cycles as constraints in an optimization problem:
	\begin{equation}
	\label{ConstrainedFormulation_PGO}
	\begin{matrix}
	\{ \mathbf{T}_{\mathbf{k}} \}_{ \mathbf{k} \in \mathcal{E} } 
	=
	\argmin\ \sum_{ \mathbf{k} \in \mathcal{E} } 
	\lVert   
	\mathbf{Log}
	(
	\mathbf{\tilde{T}}_{\mathbf{k}}^{-1} \cdot \mathbf{T}_{\mathbf{k}}
	)
	\rVert_{\boldsymbol{\Sigma}_{\mathbf{k}}}^2
	\\[5pt]
	\mathbf{s.t.}\quad
	 {\boldsymbol{\mathcal{C}}}_i^{\mathrm{lhs}} = \mathbf{I}  \quad  \forall  i \in [1 : \nu]
	.
	\end{matrix}
	\end{equation}
	This optimization problem has a degree-of-freedom (DOF) $\vert \mathcal{E} \vert- \nu = \vert \mathcal{V} \vert -1$, which is the same as the DOF of (\ref{NLLS_PGO}).
	If an optimal configuration of relative poses is found by solving (\ref{ConstrainedFormulation_PGO}),
	the objective value becomes minimum and all paths between two vertices in the graph become equivalent (guaranteed by the geometric constraints).
	Then a solution to (\ref{NLLS_PGO}) can be calculated by composing the estimates of relative poses along an arbitrary path in the graph (for example along odometry).

	In what follows, we will term the PGO formulation in (\ref{ConstrainedFormulation_PGO})
	as cycle-based PGO (\textbf{CB-PGO}),
	and in contrast, the PGO formulation in (\ref{NLLS_PGO}) as vertex-based PGO (\textbf{VB-PGO}).

\subsection{Solving Cycle Based PGO on Manifold}
	\label{Solving Edge-Cycle Based PGO by SQP}

	A typical iterative optimization algorithm on Manifold is driven by a sequence of small perturbations until convergence (to a local optimum).
	For $\mathrm{SE}(3)$, the perturbations are normally applied in the vector space of its Lie algebra,
	which can be passed to the manifold via the exponential mapping.
	To solve the PGO formulation in (\ref{ConstrainedFormulation_PGO}),
	at each iteration $t$,
	we would like to find a perturbation $\boldsymbol{\xi}_{\mathbf{k}}$ for each relative poses $\mathbf{T}_{\mathbf{k}}$,
	so that its estimate can evolve from $\hat{\mathbf{T}}_{\mathbf{k}}^{(t)}$ (estimate at iteration $t$) to $\hat{\mathbf{T}}_{\mathbf{k}}^{(t+1)}$ (estimate at iteration $t+1$) as,
	\begin{equation*}
	\hat{\mathbf{T}}_{\mathbf{k}}^{(t+1)} = \hat{\mathbf{T}}_{\mathbf{k}}^{(t)} \cdot \mathbf{Exp} \left(\boldsymbol{\xi}_{\mathbf{k}} \right)
	,\quad \mathbf{k} \in \mathcal{E}
	.
	\end{equation*}

	To this end, we linearize the PGO formulation in (\ref{ConstrainedFormulation_PGO}) with respect to the set of perturbations, to a quadratic programming with equality constraints,
		\begin{equation}
		\label{cycle-pgo linearized quadratic programming}
		\min \ \lVert \mathbf{J}^{-1} \boldsymbol{\xi} + \boldsymbol{\eta} \rVert_{\boldsymbol{\Sigma}}^2
		\qquad
		\mathbf{s.t.}\quad
		\mathbf{B} \boldsymbol{\xi} + \mathbf{b} = \mathbf{0}
		.
		\end{equation}
	Here $\mathbf{J}$ and $\boldsymbol{\eta}$ are the Jacobian matrix and the residual vector respectively by linearizing the objective function,
	whose calculations are provided in Appendix \ref{appendix cycle-pgo Linearization of Cost Function}.
	Analogously, $\mathbf{B}$ and $\mathbf{b}$ are the Jacobian matrix, and its corresponding residual vector by linearizing the geometric cycles.
	The details on how to derive $\mathbf{B}$ and $\mathbf{b}$ can be found in Appendix \ref{appendix cycle-pgo Linearization of Metric Cycles}.
	Note that the $i$-th block row and $\mathbf{k}$-th block column of $\mathbf{B}$ represents the partial derivative of the $i$-th geometric cycle with respect to the $\mathbf{k}$-th edge (i.e., relative poses),
	which is non-zero iff the $\mathbf{k}$-th edge is contained in the $i$-th cycle.
	In other words, the structure of $\mathbf{B}$ is captured by the cycle matrix $\mathcal{B}$.

	By letting
	$\bar{\boldsymbol{\xi}} = \boldsymbol{\Sigma}^{-\frac{1}{2}} 
	(\boldsymbol{\eta} + \mathbf{J}^{-1} \boldsymbol{\xi})$,
	$
	\bar{\mathbf{B}} = \mathbf{B} \mathbf{J} \boldsymbol{\boldsymbol{\Sigma}}^{\frac{1}{2}}
	$,
	and
	$
	\bar{\mathbf{b}} = \mathbf{B} \mathbf{J} \boldsymbol{\eta} - \mathbf{b}
	$,
	the quadratic programming in (\ref{cycle-pgo linearized quadratic programming}) takes the form of a minimum norm optimization problem,
		\begin{equation}
		\label{minimal norm optimzation problem}
		\min \ \lVert \bar{\boldsymbol{\xi}} \rVert^2
		\qquad
		\mathbf{s.t.}\quad
		\bar{\mathbf{B}} \bar{\boldsymbol{\xi}} = \bar{\mathbf{b}}
		\end{equation}
	whose solution is
	$
	\bar{\boldsymbol{\xi}}^{\mathrm{opt}} = \bar{\mathbf{B}}^{\dagger} \bar{\mathbf{b}}
	$.
	Note that since both $\boldsymbol{\Sigma}^{\frac{1}{2}}$ and $\mathbf{J}$ are block-diagonal matrices, $\bar{\mathbf{B}}$ and
	$\mathbf{B}$ would have the same structure.
	Finally, the perturbation in $\boldsymbol{\xi}$ can be recovered by
	$\boldsymbol{\xi}^{\mathrm{opt}} = \mathbf{J}(\boldsymbol{\Sigma}^{\frac{1}{2}} \bar{\boldsymbol{\xi}}^{\mathrm{opt}} -\boldsymbol{\eta})
	$.

	The overall algorithm can be termed as sequential quadratic programming (SQP) \cite{boggs1995sequential, bai2016incremental, bai2018robust},
	since it requires the solving of a sequence of perturbations via quadratic programming.

	\begin{remark}
		The linear system
		$\bar{\boldsymbol{\xi}}^{\mathrm{opt}} = \bar{\mathbf{B}}^{\dagger} \bar{\mathbf{b}}$ to be solved in CB-PGO has exactly the same dimension of that in the cycle space, which is $\nu = \lvert \mathcal{E} \rvert - \lvert \mathcal{V} \rvert + 1$, because the Jacobian is characterized by the cycle matrix $\mathcal{B}$.
		In contrast, an iterative solver to VB-PGO in (\ref{NLLS_PGO}) solves a linear system with a dimension of $\lvert \mathcal{V} \rvert - 1$, i.e, the dimension of cut space, since its Jacobian is characterized by the incidence matrix \cite{carlone2013convergence, carlone2014fast, khosoussi2016sparse}.
		Given the fact that the cut space and cycle space are orthogonal complementary \cite{ray2012graph},
		we conclude from the graph topology perspective that the VB-PGO and CB-PGO are counterparts to one another.
		Moreover, VB-PGO is a least squares optimization for an over-determinant system, while CB-PGO is a minimum norm optimization for an under-determinant system.
		Mathematically, the least squares optimization and minimum norm optimization are highly correlated \cite{meyer2000matrix},
		where both solutions are compactly written as Moore-Penrose pseudo inverse (i.e., left and right inverse respectively).
		This fact further confirms that VB-PGO in (\ref{NLLS_PGO}) and CB-PGO (\ref{ConstrainedFormulation_PGO}) are two sides of the same coin.
		However, while VB-PGO has reached a mature state, the CB-PGO technique is still rather primitive, because of the hardship of choosing a proper cycle basis.
		
	\end{remark}

	\subsection{Choices of Cycle Basis for PGO}

	For the cycle-based PGO in (\ref{ConstrainedFormulation_PGO}),
	the structure of the Jacobian matrices $\mathbf{B}$ and $\bar{\mathbf{B}}$ is completely described by a cycle matrix $\mathcal{B}$.
	Obviously, different choices of cycle bases $\boldsymbol{\mathcal{B}}$ lead to different cycle matrices $\mathcal{B}$, which eventually lead to different structures in $\mathbf{B}$ and $\bar{\mathbf{B}}$.

	Therefore, we discuss here the pros and cons of different cycle bases for the PGO formulation in (\ref{ConstrainedFormulation_PGO}).
	We will conclude the advantage of using a minimum cycle basis (MCB) from the sparsity perspective. The discussions on the convergence behavior will be presented in Section \ref{section: Jacobian Matrix: Invariance and MCB}.

	\subsubsection{Fundamental Cycle Basis}
	
	Given an arbitrary spanning tree of the graph,
	a cycle can be constructed by one off-tree edge (i.e. chord), and the path on the tree connecting the ends of the edge.
	The set of cycles corresponding to these $\nu$ off-tree edges are independent, called fundamental cycle basis (FCB).
	FCB can be generated cheaply, while it cannot ensure a sparse Jacobian matrix in general \cite{bai2018robust}.

	\subsubsection{Minimum Fundamental Cycle Basis}
	
	A remedy is to use the minimum fundamental cycle basis (MFCB), where the spanning tree is chosen in a way such that the summation of the lengths of the fundamental cycles is minimum.
	An exact solution to MFCB is proven to be NP-complete \cite{deo1982algorithms}.
	While approximate algorithms can solve MFCB in polynomial time
	\cite{deo1982algorithms, amaldi2009edge, galbiati2011approximability},
	constraining cycle basis to be fundamental may compromise the sparsity.

	\subsubsection{Minimum Overlap Cycle Basis}
	
	In light of the fact that the matrix to be factorized has the same structure as $\mathcal{B} \mathcal{B}^{T}$,
	the sparseness of the matrix decomposition can be guaranteed by minimizing the number of non-zeros in $\mathcal{B} \mathcal{B}^{T}$.
	We name a cycle basis that minimizes
	$\vert \mathcal{B} \mathcal{B}^{T} \vert_{0}$
	as the minimum overlap cycle basis (MOCB),
	by the fact that an entry at position $(i,j)$ in $\vert \mathcal{B} \mathcal{B}^{T} \vert_{0}$ is $0$ if and only if the cycle $i$ and $j$ do not share common edges.
	However, there is no clear way on how to compute a minimum overlap cycle basis yet.

	\subsubsection{Minimum Length Cycle Basis}
	A minimum length cycle basis (MLCB) maximizes the sparsity of $\mathcal{B}$,
	by minimizing the overall length of cycles in the basis,
	which may in turn resulting a sparse
	$\mathcal{B} \mathcal{B}^{T}$.
	Different from MFCB, the cycles are not confined to be fundamental,
	thus yielding an easier problem.
	MLCB belongs to a well-known problem termed minimum (weight) cycle basis (MCB) \cite{kavitha2009cycle}, where MLCB is a special case with weights of edges set to $1$.

	According to the discussion above,
	\textbf{we opt to use MLCB for the cycle-based PGO to maximize the sparsity}.
	Since MLCB is a special case of the general MCB, we focus on how to compute MCB in the following Section \ref{tro: section: minimum length cycle basis}.

\section{Computation of Minimum Length Cycle Basis}
\label{tro: section: minimum length cycle basis}

In this section, we aim at a complete and concise description of the MCB algorithm used for the cycle-based PGO.
We will follow a hybrid approach that firstly construct a superset that contains MCB, and then apply an independence test to extract a MCB.
We will recall the basics of these concepts for completeness,
	in particular the construction of Horton set \cite{horton1987polynomial} and isometric set \cite{amaldi2009breaking}, and the state-of-the-art method for independence tests \cite{amaldi2010efficient},
while refer interested reader to the review paper \cite{kavitha2009cycle} for further reading.

On the present architecture,
the computational bottleneck of MCB algorithms is the all-pairs-shortest-paths (APSP)
\cite{mehlhorn2007implementing, dutta2017applications}.
APSP is required to construct the Horton set, and a consistent APSP for the isometric set.
Given the fact that graphs occurred in PGO are sparse graphs with positive integer weights,
we propose two ideas that can substantially improve the performance of APSP:
1) smoothing out vertices of degree two;
2) using LexDijkstra (in Section \ref{section: mcb: lexdijkstra and parallelism}) to compute a consistent APSP that can run in parallel, and thus is more advantageous than the sequential method in \cite{hartvigsen1994all}.

The overall procedure of the proposed MCB algorithm is summarized in
Algorithm \ref{algorithm: procdure of extracting minimum cycle basis}.

\begin{algorithm}[t]
	\caption{Minimum Cycle Basis}
	\label{algorithm: procdure of extracting minimum cycle basis}
	\begin{algorithmic}
		\State{$\bar{\mathcal{G}}$ $\gets$ SimplifyGraph ($\mathcal{G}$) }
		\Comment{{\color{blue} Smooth out vertices of degree two}}
		\State{APSP $\gets$ LexDijkstra ($\bar{\mathcal{G}}$)}
		\Comment{{\color{blue} Compute a set of consistent all-pairs-shortest-paths}}
		\State{$\boldsymbol{\mathcal{H}}$ $\gets$ HortonSet (APSP)}
		\Comment{{\color{blue} Construct Horton set implicitly }}
		\State{$\boldsymbol{\mathcal{I}}$ $\gets$ IsometricSet ($\boldsymbol{\mathcal{H}}$)}
		\Comment{{\color{blue} Construct isometric set }}
		\State{$\bar{\boldsymbol{\mathcal{B}}} \gets \emptyset$}
		\Comment{{\color{blue} Initialize MCB for $\bar{\mathcal{G}}$ }}
		\State{$\boldsymbol{\mathcal{I}} \gets $ SortAscendingByWeight ($\boldsymbol{\mathcal{I}}$)}
		\State{{\color{green} // Independence test by support vectors }}
		\While{$\lvert \bar{\boldsymbol{\mathcal{B}}} \rvert \neq \nu$}
		\State{$\mathcal{C} \gets$ ExtractMinimumWeightCircuit ($\boldsymbol{\mathcal{I}}$) }
		\If{$\mathcal{C}$ is linearly independent from $\bar{\boldsymbol{\mathcal{B}}}$}
		\State{$\bar{\boldsymbol{\mathcal{B}}} \gets \bar{\boldsymbol{\mathcal{B}}} \cup \mathcal{C}$}
		\EndIf
		\State{$\boldsymbol{\mathcal{I}} \gets \boldsymbol{\mathcal{I}} \backslash \mathcal{C}$}
		\EndWhile
		\State{$\boldsymbol{\mathcal{B}} \gets $ ReconstructMCB ($\bar{\boldsymbol{\mathcal{B}}}$)}
		\Comment{{\color{blue} Reconstruct MCB for $\mathcal{G}$ }}
	\end{algorithmic}
\end{algorithm}

\subsection{Superset of MCB: Horton Set}
\label{section: mcb: Superset of MCB: Horton Set}

The study of efficient polynomial time minimum cycle basis algorithms started with Horton's work \cite{horton1987polynomial}, which builds the connection of shortest paths and a cycle in MCB.

\begin{lemma}(\cite{horton1987polynomial}).
	\label{Horton's theorem on shortest paths and cycle}
		Let $\mathcal{C}$ be a cycle in a minimum cycle basis $\boldsymbol{\mathcal{B}}$.
		If $u$ and $v$ are two vertices on $\mathcal{C}$,
		then $\mathcal{C}$ must contain one of the shortest paths from $u$ to $v$.
\end{lemma}

Another key insight from Horton \cite{horton1987polynomial} is that using shortest paths,
each cycle $\mathcal{C} \in \boldsymbol{\mathcal{B}}$ can be represented as a vertex-edge pair,
called \textsf{a representation of a cycle}.
In specific, let $x$ be any vertex in $\mathcal{C}$ ($\mathcal{C} \in \boldsymbol{\mathcal{B}}$),
then we can always find an edge $e_{uv}$
such that $\mathcal{C}$ can be expressed as
$
\mathcal{C} = \mathcal{C}(x, e_{uv}) \triangleq
\mathcal{P}_{xu} + \mathcal{P}_{xv} + e_{uv}
$,
where $\mathcal{P}_{xu}$ and $\mathcal{P}_{xv}$ are shortest paths.
Based on this observation,
Horton \cite{horton1987polynomial} proposed a superset (called \textsf{Horton set}) of MCB using all pairs of vertex-edge combinations.
Formally, a Horton set is defined as,
\begin{equation*}
\boldsymbol{\mathcal{H}} =
\{
	\mathcal{C}(x, e_{uv}) \ \vert\ 
	x \in \mathcal{V},\ e_{uv} \in \mathcal{E}
\}.
\end{equation*}

Note that there might be several shortest paths between the vertices $u$ and $v$ with exactly the same minimum weight,
so the choice of $\mathcal{P}_{uv}$ is not unique in general.
Horton \cite{horton1987polynomial} proved that
if all edges in the graph have nonnegative weights,
$\boldsymbol{\mathcal{H}}$ would definitively contain a MCB no matter what shortest path $\mathcal{P}_{uv}$ is chosen for each pair of vertices $u$ and $v$.

\begin{remark}
	Typically, $\boldsymbol{\mathcal{H}}$ is much larger than $\boldsymbol{\mathcal{B}}$.
	On the one hand,
	there are degenerated cases in $\boldsymbol{\mathcal{H}}$ that do not form a simple cycle,
	for example, if $e_{uv}$ is on $\mathcal{P}_{xu}$ or $\mathcal{P}_{xv}$,
	or if $\mathcal{P}_{xu}$ and $\mathcal{P}_{xv}$ have vertices other than $x$ in common.
	However the degenerated cases can be easily removed.
	On the other hand,
	$\boldsymbol{\mathcal{H}}$ is a multi-set.
	By choosing different vertex-edge pairs on $\mathcal{C}$, we obtain different representations of $\mathcal{C}$.
\end{remark}

\subsection{Superset of MCB: Isometric Circuits}
\label{subsection: superset of MCB: isometric circuits}

The construction of $\boldsymbol{\mathcal{H}}$ requires the computation of all pairs shortest paths (APSP).
In Horton's work \cite{horton1987polynomial},
APSP is allowed to be arbitrary,
i.e. the shortest path $\mathcal{P}_{uv}$ is selected arbitrarily among all shortest paths from $u$ to $v$.
By intentionally selecting APSP to be \textsf{consistent}, we can identify the duplicates in $\boldsymbol{\mathcal{H}}$, and reduce $\boldsymbol{\mathcal{H}}$ to a much smaller set.

\begin{definition}(Consistent APSP).
	\label{definition of consistent all-pairs-shorest-paths}
	For each shortest path $\mathcal{P}_{uv}$ in APSP, let $s$ and $t$ be two arbitrary vertices lying on $\mathcal{P}_{uv}$, then $\mathcal{P}_{st}$, the selected shortest path from $s$ to $t$ in APSP, is a subgraph of $\mathcal{P}_{uv}$.
\end{definition}

A consistent APSP can be computed by a lexicographic method \cite{hartvigsen1994all, kavitha2009cycle}
(see Definition \ref{appendix def lexicographic ordering by edge ids} and Lemma \ref{appendix theorem: lexicographic paths to consistent APSP}
in Appendix \ref{appendix. consistent APSP. lexicographic djikstra}).
Given a consistent APSP, a cycle $\mathcal{C}$ is said to be \textsf{isometric} if for any two vertices $u$ and $v$ on $\mathcal{C}$, the chosen shortest path $\mathcal{P}_{uv}$ in APSP is contained in $\mathcal{C}$ \cite{horton1987polynomial, amaldi2009breaking}.
It can be further verified that a cycle $\mathcal{C}$ is isometric if and only if, for each vertex $x \in \mathcal{C}$,
there is a unique edge $e_{uv}$, such that $\mathcal{C} = \mathcal{C}(x, e_{uv})$,
with $\mathcal{P}_{xu}$ and $\mathcal{P}_{xv}$ being in the consistent APSP
\cite{amaldi2009breaking}.
Last but not least, the set of all isometric cycles is proved to contain a MCB \cite{kavitha2009cycle, amaldi2009breaking}.

In a Horton set $\boldsymbol{\mathcal{H}}$ constructed by a consistent APSP,
an isometric cycle $\mathcal{C} \in \boldsymbol{\mathcal{H}}$ would contain exactly $\lvert \mathcal{C} \rvert$ representations, i.e. $\lvert \mathcal{C} \rvert$ duplicates in $\boldsymbol{\mathcal{H}}$.
Aiming to eliminate redundant representations, 
we will use the following Lemma to find all the equivalent representations in the Horton set.

\begin{lemma}(\cite{amaldi2009breaking}).	
	\label{lemma isometric circuit identification}
Let $s_x(y)$ be the first vertex (except $x$) on the shortest path $\mathcal{P}_{xy}$.
For any cycle $\mathcal{C} = \mathcal{C}(x, e_{uv}) \in \boldsymbol{\mathcal{H}}$,
with $e_{uv} \notin \mathcal{P}_{xu}$,
$e_{uv} \notin \mathcal{P}_{xv}$,
and $s_x(u) \neq s_x(v)$,
\begin{enumerate}
\item[(1)] if $x=u$ then $\mathcal{C} = \mathcal{C} (v, e_{uv})$.
\item[(2)] if $x \neq u$, let $x' = s_x(u)$,
	\begin{itemize}
		\item[(a)] if $x = s_{x'}(v)$ then $\mathcal{C} = \mathcal{C}(x',e_{uv})$.
		\item[(b)] if $x \neq s_{x'}(v)$, $u = s_v(x')$ then $\mathcal{C} = \mathcal{C}(v, e_{xx'})$.
		\item[(c)] if $x \neq s_{x'}(v)$, $u \neq s_v(x')$ then $\mathcal{C}$ is not isometric.
	\end{itemize}
\end{enumerate}
\end{lemma}

\begin{figure}[t]
	\centering
	\begin{tikzpicture}
	
	\def \vldist{0.6}
	\def \dx{4.5}
	\def \dy{2.0}

	\tikzset{VertexStyle1/.style = {
			shape = circle,
			draw=black!80, 
			fill= cyan!60,
			very thick,
			solid,
			inner sep      = 2pt,
			outer sep      = 2pt,
			minimum size   = 24 pt}}
	
	\tikzset{VertexStyle2/.style = {
			shape = ellipse,
			draw= black!80, 
			fill= green!60,
			very thick,
			solid,
			inner sep      = 2pt,
			outer sep      = 2pt,
			minimum size   = 24 pt}}

	\tikzset{EdgeStyle1/.style   = {very thick,
			blue!80,
		}}
		
	\tikzset{EdgeStyle2/.style   = {very thick,
			dashed,
			red!80
		}}

			\node[VertexStyle1] at (0,0) (a) {$x$};
			
			\node[VertexStyle2] at (0,\dy) (b) {$x'= s_x(u)$};
			
			\node[VertexStyle1] at (\dx,\dy) (d) {$u$};
			
			\node[VertexStyle1] at (\dx, 0) (e) {$v$};

			\draw[EdgeStyle1] (a) -- (b) node[midway, label=left:{\color{black}$e_{xx'}$}]{};	    
			\draw[EdgeStyle2] (a) -- (e) node[midway, label=above:{\color{black}$\mathcal{P}(x, v)$}]{};	    
			\draw[EdgeStyle1] (d) -- (e) node[midway, label=right:{\color{black}$e_{uv}$}]{};	    
			\draw[EdgeStyle2] (b) -- (d) node[midway, label=below:{\color{black}$\mathcal{P}(x', u)$}]{};
			
			\end{tikzpicture}
			\caption{An illustration of Lemma \ref{lemma isometric circuit identification}.
				Let us consider the cycle represented by the vertex $x$ and edge $e_{uv}$, i.e.
				$
				\mathcal{C} = \mathcal{C}(x, e_{uv}) \triangleq
				\mathcal{P}(x, u) + \mathcal{P}(x, v) + e_{uv}
				$.
				$x' = s_x(u)$ is the first vertex on the path $\mathcal{P}(x, u)$, i.e., $\mathcal{P}(x, u) = e_{xx'} + \mathcal{P}(x',u)$.
				The key to the proof of Lemma \ref{lemma isometric circuit identification} is based on the observation that: 
				If $\mathcal{C}$ is isometric,
				there are two possible cases for the shortest path between the vertex $x'$ and $v$:
				(a) $\mathcal{P}(x', v) = e_{xx'} + \mathcal{P}(x,v) $,
				(b) $\mathcal{P}(x', v) = \mathcal{P}(x',u) + e_{uv} $.
				Obviously case (a) implies $x = s_{x'}(v)$ and $\mathcal{C} = \mathcal{C}(x',e_{uv})$,
				while case (b) implies $u = s_v(x')$ and $\mathcal{C} = \mathcal{C}(v, e_{xx'})$.}
			\label{fig: an intuive example of Lemma on isometric circuits}	
		\end{figure}
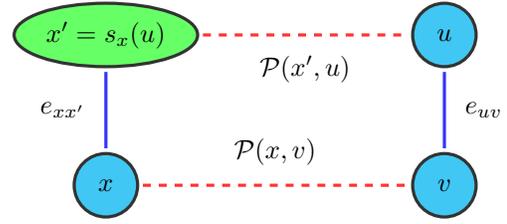

\begin{IEEEproof}
	The proof can be found in \cite{amaldi2009breaking}.
	Note that the cases $e_{uv} \in \mathcal{P}_{xu}$,
	$e_{uv} \in \mathcal{P}_{xv}$,
	and $s_x(u) = s_x(v)$
	create bridges, thus do not form cycles and need to be excluded.
	An intuitive explanation of isometric cases is presented in Fig. \ref{fig: an intuive example of Lemma on isometric circuits}.
\end{IEEEproof}

	In Lemma \ref{fig: an intuive example of Lemma on isometric circuits}
	and Fig. \ref{fig: an intuive example of Lemma on isometric circuits},
	$x'$ is chosen on the path $\mathcal{P}(xu)$.
	However, we can also chose $x'$ on the path $\mathcal{P}(xv)$, i.e., letting $x'= s_x(v)$.
	By doing so, we can find another equivalent representation for an isometric circuit $\mathcal{C}$ by Lemma \ref{fig: an intuive example of Lemma on isometric circuits}.

The connection of different representations can be visualized as a directed graph $\mathcal{G}^{\dagger} = \mathcal{G}^{\dagger} (\mathcal{V}^{\dagger}, \mathcal{E}^{\dagger})$,
where each vertex in $\mathcal{G}^{\dagger}$ corresponds to a cycle in the Horton set.
If a cycle $\mathcal{C}(x, e_{uv})$ is equivalent to a cycle $\mathcal{C}(y, e_{st})$ by Lemma \ref{lemma isometric circuit identification},
then an arc is formed from the vertex $\mathcal{C}(x, e_{uv})$ to the vertex $\mathcal{C}(y, e_{st})$ in $\mathcal{G}^{\dagger}$.
It was proved that all representations of an isometric cycle $\mathcal{C}$ exactly correspond to a single connected component in $\mathcal{G}^{\dagger}$ \cite{amaldi2009breaking}.
The below theorem will greatly improve the efficiency of operations on $\mathcal{G}^{\dagger}$, in terms of graph storage and searching.

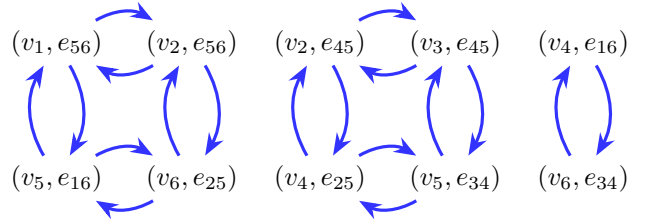
\begin{figure}[t]
	\centering
	\begin{tikzpicture}
	
	\def \dx{3.5}
	\def \dy{1.8}
	
	\tikzset{VertexStyle1/.style = {
			shape = circle,
			draw=black!80, 
			fill= cyan!60,
			very thick,
			solid,
			inner sep      = 2pt,
			outer sep      = 2pt,
			minimum size   = 20 pt}}

	\tikzset{EdgeStyle1/.style   = {very thick,
			blue!80,
			->,
			>=Stealth,
		}}
		\tikzset{EdgeStyle2/.style   = {very thick,
				blue!80,
				->,
				>=Stealth,
			}}

			\node[] at (0,0) (a) {$(v_5, e_{16})$};
			\node[] at (0,\dy) (b) {$(v_1, e_{56})$};
			\node[] at (\dy,\dy) (c) {$(v_2, e_{56})$};
			\node[] at (\dy, 0) (d) {$(v_6, e_{25})$};
			
			\draw[EdgeStyle1] (a) to[bend left] (b);
			\draw[EdgeStyle2] (b) to[bend left] (a);	
			\draw[EdgeStyle1] (b) to[bend left] (c);
			\draw[EdgeStyle2] (c) to[bend left] (b);
			\draw[EdgeStyle1] (c) to[bend left] (d);
			\draw[EdgeStyle2] (d) to[bend left] (c);
			\draw[EdgeStyle1] (a) to[bend left] (d);
			\draw[EdgeStyle2] (d) to[bend left] (a);

			\node[] at (\dx,0) (e) {$(v_4, e_{25})$};
			\node[] at (\dx,\dy) (f) {$(v_2, e_{45})$};
			\node[] at (\dx+\dy,\dy) (g) {$(v_3, e_{45})$};
			\node[] at (\dx+\dy, 0) (h) {$(v_5, e_{34})$};

			\draw[EdgeStyle1] (e) to[bend left] (f);
			\draw[EdgeStyle1] (f) to[bend left] (e);	
			\draw[EdgeStyle1] (f) to[bend left] (g);
			\draw[EdgeStyle1] (g) to[bend left] (f);
			\draw[EdgeStyle1] (g) to[bend left] (h);
			\draw[EdgeStyle1] (h) to[bend left] (g);
			\draw[EdgeStyle1] (h) to[bend left] (e);
			\draw[EdgeStyle1] (e) to[bend left] (h);

			\node[] at (2*\dx,\dy)(i) {$(v_4, e_{16})$};
			\node[] at (2*\dx, 0) (j) {$(v_6, e_{34})$};
			
			\draw[EdgeStyle1] (i) to[bend left] (j);
			\draw[EdgeStyle2] (j) to[bend left] (i);
			
	\end{tikzpicture}
	\caption{A visualization of connected components in $\mathcal{G}^{\dagger}$.
	Each isometric circuit $\mathcal{C}$ in $\mathcal{G}$ corresponds to a double-linked directed cycle in $\mathcal{G}^{\dagger}$ with $\lvert \mathcal{C} \rvert$ vertices.
	This example is created using the graph in Fig. \ref{fig An illustration of incidence matrix and cycle matrix of a graph}, while the cycle representations that do not create links by Lemma \ref{lemma isometric circuit identification} are ignored.}
	\label{fig: an intuive example of Thereom 1 on double linked directed cycle}
\end{figure}

\begin{theorem}
	\label{thoerem: property of connected component}
	All representations of an isometric circuit $\mathcal{C}$ in $\mathcal{G}^{\dagger}$ form a double-linked directed cycle with $\vert \mathcal{C} \vert$ vertices. 
\end{theorem}
\begin{IEEEproof}
	See Appendix \ref{proof of Theorem on the property of connected component}.
	A visualization of connected components in $\mathcal{G}^{\dagger}$ is given in Fig. \ref{fig: an intuive example of Thereom 1 on double linked directed cycle}.
\end{IEEEproof}

Based on Theorem \ref{thoerem: property of connected component}, the storage of $\mathcal{G}^{\dagger}$ can be compressed to a vector with $2$ slots reserved for each vertex. We can easily access the adjacent vertices of a given vertex by its index.
Besides, a depth-first-search (DFS) \cite{ray2012graph} on connected components can be simplified as: starting from an arbitrary vertex, keep exploring new vertices until coming back to the start-vertex.
This eliminates the use of function recursions or data structures like a stack.

Finally, we characterize the connected components corresponding to isometric cycles by the above simplified DFS.
Duplicate representations of an isometric cycle are then removed by keeping only one representation in the connected component.
All representations of non-isometric cycles are discarded.
The construction of isometric cycles from a Horton set $\boldsymbol{\mathcal{H}}$ (constructed by a consistent APSP) can be achieved with an amortized complexity $O (\lvert \mathcal{V} \rvert \lvert \mathcal{E} \rvert)$ \cite{amaldi2009breaking}.

\begin{remark}
	It should be noted that the set of all isometric cycles is a superset to a MCB, but not all MCBs.
	In other words, even though APSP is chosen to be consistent,
	a cycle $\mathcal{C}$ in an arbitrary MCB can be non-isometric \cite{kavitha2009cycle}.
\end{remark}

\subsection{Independence Test}
\label{sec: mcb: independecne test}

To extract a MCB, we sort the set of isometric cycles in non-descending order of weights,
and sequentially extract $\nu$ linearly independent cycles with the least weights.
This procedure is proved to find a MCB \cite{horton1987polynomial}.
To test the linear independence,
we take a circuit as a vector on $\mathbb{Z}_2^m$ incident on the set of edges,
and then evaluate the linear independence algebraically.

Given a (spanning) tree $\mathcal{T}$, let the \textsf{restricted incidence vector} of $\mathcal{C}$ be $\bar{\mathcal{C}} = \mathcal{C} \backslash \mathcal{T}$, i.e. considering the off-tree edges of $\mathcal{T}$ only \cite{de1995applications}.
It can be shown that the linear independence of a collection of cycles $\{ \mathcal{C}_i \}_{i \in N}$ implies the linear independence of the corresponding restricted incidence vectors $\{ \bar{\mathcal{C}}_i \}_{i \in N}$,
and vice versa
(see Theorem \ref{Theorem: rank relation restricted and full cycle incidence} in Appendix \ref{Appendix Proof of Theorem on graph theory} for a proof).

\subsubsection{ Gaussian Elimination Based Approach }
Gaussian elimination is a well-exploited technique in graph theory to check the linear independence of incidence vectors \cite{horton1987polynomial, golynski2002polynomial}.
The basic idea is to stack the set of (restricted) incidence vectors as a matrix which will be subsequently reduced to the row echelon form.
The incidence vectors are linearly independent if and only if the row echelon form has full row rank.
This approach has a cubic complexity in the worst case.

\subsubsection{ Support Vector Based Approach }

Let $\{ \mathcal{C}_i \}_{i=1}^{k-1}$ be a set of independent circuits.
Given a spanning tree $\mathcal{T}$,
let $\{ \bar{\mathcal{C}}_i \}_{i=1}^{k-1}$ be the corresponding restricted incidence vectors
whose span is a space $\mathbf{C}_{ [1:{k-1}] }$.
Denote the orthogonal complementary space as $\mathbf{S}_{ [k:{\nu}] } = \mathbf{C}_{ [1:{k-1}] }^{\perp}$.
Let $\{ \bar{\mathcal{S}}_i \}_{j=k}^{\nu}$ be a basis of the space $\mathbf{S}_{ [k:{\nu}] }$.
The vectors $\{ \bar{\mathcal{S}}_i \}_{j=k}^{\nu}$ are called \textsf{support vectors} of $\{ \mathcal{C}_i \}_{i=1}^{k-1}$ \cite{kavitha2008tilde, kavitha2009cycle}.
	Then a circuit $\mathcal{C}_k$ is linearly independent from the set of independent circuits $\{ \mathcal{C}_i \}_{i=1}^{k-1}$,
	if and only if there exists a $\bar{\mathcal{S}}_l$ $(k \le l \le \nu)$, such that $\langle \bar{\mathcal{C}}_k, \bar{\mathcal{S}}_l \rangle = 1$ (see Lemma \ref{lemma: independence of cycles based on support vectors} in Appendix \ref{Appendix Proof of Theorem on graph theory}).
	If such a $\bar{\mathcal{S}}_l$ is found, then $\{ \mathcal{C}_i \}_{i=1}^{k} = \{ \mathcal{C}_i \}_{i=1}^{k-1} \cup \mathcal{C}_k$ are a set of independent circuits.
	The support vectors of $\{ \mathcal{C}_i \}_{i=1}^{k}$, i.e. a basis of the space
	$
	\mathbf{S}_{ [k+1:{\nu}] } = \mathbf{C}_{ [1:{k}] }^{\perp}
	$
	can be obtained by updating $\{ \bar{\mathcal{S}}_i \}_{j=k, j \ne l}^{\nu}$ using $\bar{\mathcal{S}}_l$ \cite{kavitha2008tilde, kavitha2009cycle}.
	Let
		\begin{equation*}
		\bar{\mathcal{S}}'_j =
		\begin{cases}
		\bar{\mathcal{S}}_j   & \qquad \mathrm{if}\ \langle \bar{\mathcal{C}}_k, \bar{\mathcal{S}}_l \rangle = 0  \\
		\bar{\mathcal{S}}_j + \bar{\mathcal{S}}_l  & \qquad   \mathrm{if}\ \langle \bar{\mathcal{C}}_k, \bar{\mathcal{S}}_l \rangle = 1 \\
		\end{cases}
		,
		\end{equation*}
	then $ \{ \bar{\mathcal{S}}'_j \}_{j=k, j \ne l}^{\nu}$ 
	is a set of support vectors for $\{ \mathcal{C}_i \}_{i=1}^{k}$.

	A direct use of support vectors to check independence can be found in \cite{mehlhorn2007implementing}.
	We invert the process in \cite{mehlhorn2007implementing} to accommodate our description.
	Given a spanning tree $\mathcal{T}$, we can initialize $\nu$ independent support vectors by $\nu$ off-tree edges, where each support vector contains one off-tree edge \cite{mehlhorn2007implementing}.
	Then at each phase, we evaluate the independence of a new circuit $\mathcal{C}$ by support vectors, which will be subsequently updated if $\mathcal{C}$ is evidenced to be independent by a support vector $\mathcal{S}_l$.
	Iterate this process until a MCB is found.
	As in \cite{mehlhorn2007implementing}, the drawback of this method is that we might need to check many support vectors in order to verify $\langle \mathcal{C},\mathcal{S}_l \rangle = 1$.

	A more sophisticated design is due to Amaldi et al. \cite{amaldi2010efficient}.
	The approach is	based on the idea that if a circuit $\mathcal{C}$ contains edges that are not used by any selected circuits, then this circuit is independent from the selected ones.
	If this is the case, we can verify the independence of $\mathcal{C}$ without using any support vector.
	Moreover, the ``new" edges in $\mathcal{C}$ can be used to construct new support vectors.
	Particularly,
	there is no need to designate a spanning tree $\mathcal{T}$ and initialize a set of independent support vectors at the beginning of the algorithm.
	The spanning tree $\mathcal{T}$ is built adaptively by greedily including ``new" edges without creating a cycle to maximize the sparsity of $\mathcal{C}$.
	Let the new edges in $\mathcal{C} \backslash \mathcal{T}$ be $\mathcal{C}_N = \{ e_1, \dots, e_k \}$, then we can identify maximally $k$ independent support vectors,
	for example,
	$\mathcal{S}_0 = \{e_1\}$ and $\mathcal{S}_j = \{ e_j, e_{j+1}\}\ (1 \le j \le k-1)$.
	Obviously, $\langle \mathcal{C}, \mathcal{S}_0 \rangle = 1$ and $\langle \mathcal{C}, \mathcal{S}_k \rangle = 0$,
	which means $\mathcal{S}_0$ is an implicit support vector that evidences $\mathcal{C}$,
	and there is no need to update $\mathcal{S}_k$ for $\mathcal{C}$.
	If $\mathcal{C}$ does not contain any new edge, the algorithm checks the existing support vectors by Lemma \ref{lemma: independence of cycles based on support vectors} to verify the independence instead.
	To speed up the inner product $\langle \mathcal{C}, \mathcal{S} \rangle$,
	the algorithm maintains
	$\mathcal{E}_{\mathcal{S}}$ to be the edges used by present support vectors,
	and $\mathcal{E}^{\circ}$ to be those not anymore because of the update of support vectors
	($\mathcal{E}_{\mathcal{S}} \cup \mathcal{E}^{\circ}$ is the set of off-tree edges).
	At each stage, the edges in both $\mathcal{T}$ and $\mathcal{E}^{\circ}$ can be excluded to increase the sparsity of $\mathcal{C}$.

We will use the algorithm by Amaldi et al. \cite{amaldi2010efficient} to extract a MCB from the set of isometric cycles.

\subsection{Smoothing Out Vertices of Degree Two}
\label{section: Smoothing Out Nodes of Degree Two}

For PGO, the underlying graph is usually sparse.
\textbf{Furthermore, we assume that the sparsity of PGO is positively correlated to the proportion of vertices of degree two in the graph.}
The sparser the graph is, the more vertices of degree two we have.
The vertices of degree two have no contribution to the topology of the graph, thus can be pruned out for the computation of a MCB.
The pruning of vertices of degree two, along with the edges incident to them,
would greatly reduce the combinatorial complexity of the Horton set.

\begin{figure}
	\centering
	\subfloat[Original unweighted graph.]
	{	\begin{tikzpicture}
		
		\def \xgridlen{1.5}
		\def \rotangle{60} 
		\def \ygridlen{1.4 * sin{\rotangle}}
		\def \dx{\xgridlen * cos{\rotangle}}

		\tikzset{VertexStyle1/.style = {
				shape = circle,
				draw=black!80, 
				fill= cyan!60,
				thick,
				solid,
				minimum size   = 15 pt }}

		\tikzset{VertexStyle2/.style = {
				shape = circle,
				draw= red!80, 
				fill= red!40,
				thick,
				solid,
				minimum size   = 15 pt }}
		
		\tikzset{EdgeStyle1/.style   = {
				line width=1.2pt }}
		
		\tikzset{NodeStyle1/.style = {
				thick,
				solid,
				minimum size   = 5 pt }}

		\node[VertexStyle1] at (\xgridlen, 0) (v1) {$v_1$};
		\node[VertexStyle1] at (2*\xgridlen, 0) (v2) {$v_2$};
		\node[VertexStyle1] at (3*\xgridlen, 0) (v3) {$v_3$};
		\node[VertexStyle2] at (4*\xgridlen, 0) (v4) {$v_4$};
		\node[VertexStyle2] at (4*\xgridlen + \dx, -\ygridlen) (v5) {$v_5$};
		\node[VertexStyle2] at (3*\xgridlen + \dx, -\ygridlen) (v6) {$v_6$};
		\node[VertexStyle2] at (3*\xgridlen - \dx, -\ygridlen) (v7) {$v_7$};		
		\node[VertexStyle2] at (2*\xgridlen - \dx, -\ygridlen) (v8) {$v_8$};

		\draw[EdgeStyle1] (v1) -- (v2);
		\draw[EdgeStyle1] (v2) -- (v3);
		\draw[EdgeStyle1] (v3) -- (v4);
		\draw[EdgeStyle1] (v4) -- (v5);
		\draw[EdgeStyle1] (v5) -- (v6);
		\draw[EdgeStyle1] (v6) -- (v3);
		\draw[EdgeStyle1] (v3) -- (v7);
		\draw[EdgeStyle1] (v7) -- (v8);
		\draw[EdgeStyle1] (v8) -- (v2);

		\end{tikzpicture}
	}
	\\
	\subfloat[Reduced weighted graph.]
	{	
		\begin{tikzpicture}
	
		\def \xgridlen{1.7}
		\def \rotangle{75} 
		\def \ygridlen{\xgridlen * sin{\rotangle}}
		\def \dx{\xgridlen * cos{\rotangle}}

		\tikzset{VertexStyle1/.style = {
				shape = circle,
				draw=black!80, 
				fill= cyan!60,
				thick,
				solid,
				minimum size   = 20 pt}}
		
		\tikzset{EdgeStyle1/.style   = {
				line width=1.2pt }}
		
		\tikzset{NodeStyle1/.style = {
				thick,
				solid,
				minimum size   = 5 pt}}

		\node[VertexStyle1] at (\xgridlen, 0) (v1) {$v_1$};
		\node[VertexStyle1] at (2*\xgridlen, 0) (v2) {$v_2$};
		\node[VertexStyle1] at (3*\xgridlen, 0) (v3) {$v_3$};

		\draw[EdgeStyle1] (v1) -- (v2) node[NodeStyle1, midway, above]{$1$};
		\draw[EdgeStyle1] (v2) -- (v3) node[NodeStyle1, midway, above]{$1$};
	
		\path (v2) 
			edge[EdgeStyle1, out= -50, in= -130]
			node[NodeStyle1, midway, above]{$3$}
			(v3);

		\path[loop/.style={looseness=18}] 	(v3) 
			edge[EdgeStyle1, in= 20,out= -20, loop]
			node[NodeStyle1, midway, right]{$4$}
			(v3);
			
		\end{tikzpicture}
	}
	\caption{The original graph (unweighted) and the corresponding reduced graph (weighted) by smoothing out the vertices of degree two. The weight of an edge in the reduced graph is the number of edges it represents in the original graph. Both graphs possess the same cycle structure.}
	\label{fig. A figure on smoothing out nodes of degree two}
\end{figure}
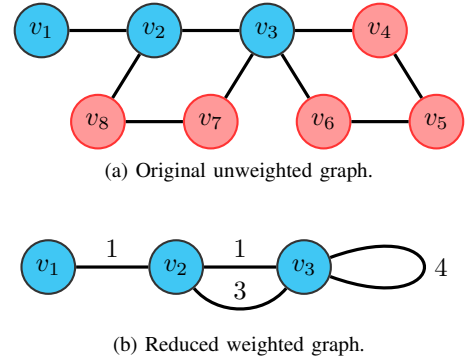

Algorithmically, we can perform a DFS from a node whose degree is not two.
During the search, if a vertex of degree two is detected,
we greedily probe along the ``degree two chain" until a vertex whose degree is not two is found.
Then we replace the ``degree two chain" by a new edge (let us name it a ``chain edge") whose weight is the accumulated weight along the chain (see Fig. \ref{fig. A figure on smoothing out nodes of degree two}).
The DFS is recursively called at every unvisited vertex whose degree is not two.
Finally, after computing a MCB on the reduced graph, the ``chain edges" can be replaced back by the corresponding ``degree two chains"
to obtain a MCB of the original graph.

\begin{remark}(Ear Decomposition).
The vertices of degree two can also be pruned out using the ear decomposition \cite{ramachandran1992parallel},
which requires the graph to be  2-edge connected.
%
%
A minimum cycle basis algorithm exploiting ear decomposition is provided in \cite{dutta2017applications}.
The algorithm in \cite{dutta2017applications} exploiting the idea of feedback vertices to reduce the Horton set, which is encompassed in the concept of isometric cycles \cite{amaldi2010efficient}.
Nevertheless, the set of isometric cycles are much smaller than the set of cycles exploiting the idea of feedback vertices \cite{amaldi2010efficient}.
Actually, an ear decomposition is closely related to a depth-first-search (DFS) of the graph \cite{ramachandran1992parallel}.
As a result, the task of pruning vertices of degree two can be achieved by DFS explicitly without computing an ear decomposition as an intermediate step.

\end{remark}

\subsection{Self Loops and Multiple Edges}

Sometimes the graph may contain self-loops or multiple-edges
which can be created artificially, or as a consequence of smoothing out vertices of degree two (see Section \ref{section: Smoothing Out Nodes of Degree Two} and an example in Fig. \ref{fig. A figure on smoothing out nodes of degree two}).
The self-loops and multiple-edges can be easily coped with by describing paths and cycles as a set of edges (instead of vertices) in the MCB algorithm.
It is also possible to eliminate all self-loops and multiple-edges for the MCB computation (see Lemma 3.17 and Lemma 3.18 in \cite{kavitha2009cycle}).
However, as shown in the experiments, the computational bottle-neck of the MCB is APSP, while the cost on the independence test is negligible.
Thus we retain self-loops and multiple-edges in the graph, and opt to
use edges to describe paths and cycles.

\subsection{LexDijkstra and Parallelism}
\label{section: mcb: lexdijkstra and parallelism}

The bottleneck of the described MCB algorithm is the computation of a consistent APSP (see Table \ref{table: LexDijkstra and MCB timing validation on benchmarks.}).

The method described in \cite{hartvigsen1994all} first compute all-pairs-shortest-distances (APSD) (by using any shortest paths algorithm),
then a set of consistent APSP is constructed by choosing the so-called lexicographic shortest path for each pair of vertices
(see Definition \ref{appendix def lexicographic ordering by edge ids} and Lemma \ref{appendix theorem: lexicographic paths to consistent APSP} in Appendix \ref{appendix. consistent APSP. lexicographic djikstra}),
by processing vertex-pairs according to distances (i.e. weight and length) from the shortest to the longest.
Obviously, a sorting process is required \cite{hartvigsen1994all},
which can be mitigated by a topological sorting \cite{kavitha2009cycle}.
Besides, although not shown in amortized complexity,
the random access to shortest path trees is expensive,
in particular for a serial processing (by distances).

The algorithm in \cite{hartvigsen1994all} is general, applicable for graphs with negative weights and any APSP algorithm.
However, for a sparse graph with positive weight, Dijkstra \cite{dijkstra1959note} is always the preferable shortest paths algorithm.
It can be shown that the lexicographic shortest path for each vertex-pairs can be selected by slightly modifying Dijkstra's update process.
%
The resultant APSP is to run Dijkstra for each vertex, which can be easily parallelized by using a multi-core CPU.
We refer to this consistent APSP method as \textbf{LexDijkstra}.

		\begin{proposition}(LexDijkstra).
			\label{appendix. proposition Lexicographic Dijkstra}
			Let $\mathcal{G}(\mathcal{V}, \mathcal{E})$ be an undirected graph with weight $w(e) > 0, \forall e \in \mathcal{E}$.
			A consistent shortest path $\mathcal{P}_{uv}$ can be obtained for each pair of nodes $u$, $v$ by modifying Dijkstra's update process to choose the lexicographic path in Definition \ref{appendix def lexicographic ordering by edge ids} (provided
				in Appendix \ref{appendix. consistent APSP. lexicographic djikstra}).
		\end{proposition}
	\begin{IEEEproof}
		See Appendix \ref{proof of proposition: LexDijkstra} for the proof, and Theorem \ref{appendix theorem Dijkstra property - all paths serached} in Appendix \ref{appendix. consistent APSP. lexicographic djikstra} that supports this result.
	\end{IEEEproof}

In terms of the lexicographic comparison defined in Definition \ref{appendix def lexicographic ordering by edge ids}, case (1) and case (2) are rather cheap.
However, case (3) in the worst case, needs to traverse and compare all the edges in $\mathcal{P}'_{uv}$ and $\mathcal{P}'_{uv}$, which is rather inefficient.
To address this issue, we propose the following theorem that greatly reduces the complexity in case (3).

\begin{theorem}
	\label{theorem: LexDijkstra, only need to traverse back to a common node.}
	Let $\mathcal{P}_{uv}$ and $\mathcal{P}'_{uv}$ be two paths from $u$ to $v$.
	In LexDijkstra, if the algorithm reaches the case (3) of Definition \ref{appendix def lexicographic ordering by edge ids},
	it just suffices that the algorithm traverses back to the nearest common vertex that shared by $\mathcal{P}_{uv}$ and $\mathcal{P}'_{uv}$.
\end{theorem}

\begin{IEEEproof}
	See Appendix \ref{proof of theorem on traverse back to a common node}.
\end{IEEEproof}

For the construction of isometric circuits, the algorithm requires random access to the shortest path trees. However, this part can be parallelized without any additional effort.

\subsection{Complexity}

The overall computational time of the proposed MCB algorithm is presented in Table \ref{table: LexDijkstra and MCB timing validation on benchmarks.}, running on a quad-core CPU.
Table \ref{table: LexDijkstra and MCB timing validation on benchmarks.} shows that the time spending on the independence test is negligible compared with that spending on computing a consistent APSP and constructing the isometric set.
Let $ \bar{\mathcal{G}} (\bar{\mathcal{V}},  \bar{\mathcal{E}})$ be the reduced graph.
Let $m = \lvert \bar{\mathcal{E}}  \rvert$, $n = \lvert \bar{\mathcal{V}}  \rvert$.

\subsubsection{Computing a Consistent APSP}
Let us compare the proposed LexDijkstra with the Method in \cite{hartvigsen1994all}.
\textsf{LexDijkstra:}
The Dijkstra algorithm has a sorting bottleneck which is usually addressed by a priority queue.
In one single Dijkstra run,
for each new edge,
the operation on the priority queue has a complexity $O(\log n)$.
In the worst case, the lexicographic comparison between two paths takes $O(n)$ operations.
Therefore, each new edge contributes
$O(\log n + n)$ worst case complexity,
which results in
$O(m (\log n + n))$
operations for one single LexDijkstra run,
and
$O(n m (\log n + n)$
operations to compute a consistent APSP using LexDijkstra.
\textsf{Method in} \cite{hartvigsen1994all}:
By the method in \cite{hartvigsen1994all},
the overall operations used to compute a consistent APSP is $O (n m \log n +
n^2 \log n
 + n m )$,
where the term $O(n m \log n)$ accounts for operations to compute an arbitrary APSP based on the Dijkstra algorithm,
$O(n^2  \log n)$ for sorting paths according to weights and lengths,
and $O (n m)$ for constructing the consistent shortest paths.

By the worst cases complexity, it seems that LexDijkstra does not offer any benefits compared with the method in \cite{hartvigsen1994all}.
However, this is not the cases in practice (see Table \ref{table: LexDijkstra and MCB timing validation on benchmarks.}).
The reason is four folds:
First, $O (n m)$ in the method \cite{hartvigsen1994all} cannot be eliminated because it requires random access to all pairs shortest path trees, which is expensive on modern architectures.
Second, 
the method in \cite{hartvigsen1994all} cannot be run in full parallelism since the sorting
term $O(n^2  \log n)$
and processing term $O (n m)$ are sequential operations.
Third, $O(n)$ is the worst case complexity for lexicographic comparison,
while in practice the operations can be greatly reduced due to the inherent asymmetry in the graph and by Theorem \ref{theorem: LexDijkstra, only need to traverse back to a common node.} as well.
Last but not least, LexDijkstra can run in full parallelism which can take advantage of multi-core CPU architectures.

\subsubsection{Constructing the Isometric Set}
It takes
$O (n m)$ operations to find a single representation for each isometric circuit \cite{amaldi2009breaking}.
However there is a big constant due to the random access to the all pairs shortest path trees,
which is implemented as a dense $n \times n$ matrix.
While the running time of this part is not major in Table \ref{table: LexDijkstra and MCB timing validation on benchmarks.} compared with the expenses on the consistent APSP,
we believe this part can be further improved using a sparse storage for all pairs shortest path trees,
given that most of the elements in the dense matrix are redundant because of the consistency of shortest paths.

\section{Discussions}
\label{section: analyses on cycle based PGO}

\subsection{Observability}

We propose to define the observability in the maximum likelihood estimation (MLE) as:
\begin{definition}(Observability of MLE \cite{jauffret2007observability})
	\label{defintion of observablity in maximum likelihood estimation problem}
	Let $\mathcal{M}^{d}$ be a manifold of dimension $d$.
	A noise-free system $\mathbf{Z} = {\mathfrak{F}}(\mathbf{X}) : \mathcal{M}_1^{d_1} \rightarrowtail \mathcal{M}_2^{d_2}$,
	is locally observable at $\mathbf{X}_0 \in \mathcal{M}_1^{d_1}$ if there is a neighborhood of $\mathbf{X}_0$, denoted by $\mathbb{U}_{\mathbf{X}_0}$, such that
	\begin{equation*}
		\forall \mathbf{X} \in \mathbb{U}_{\mathbf{X}_0},
		\ \mathbf{X} \neq \mathbf{X}_0,
		\ \mathrm{we\ have}\ 
		{\mathfrak{F}}(\mathbf{X}) \neq {\mathfrak{F}}(\mathbf{X}_0)
		.
	\end{equation*}
\end{definition}

By Definition \ref{defintion of observablity in maximum likelihood estimation problem},
it is easy to verify that VB-PGO in (\ref{NLLS_PGO}) is unobservable.
Because given a solution $\mathbf{X}_0 \triangleq \{ \mathbf{T}_i \}_{i \in \mathcal{V}}$
and any neighborhood $\mathbb{U}_{\mathbf{X}_0}$ around $\mathbf{X}_0$,
by the fact that a Lie group is a continuous group,
we can always find a shifted solution
$\mathbf{X} \triangleq \{ \mathbf{T}' \mathbf{T}_i \}_{i \in \mathcal{V}}$,
$\mathbf{X} \in \mathbb{U}_{\mathbf{X}_0}$,
which yields exactly the same measurements
(see Remark \ref{remark traditional pgo is unobservable}).

This result coincides with the Fisher information matrix (FIM) based observability tool
(see Remark \ref{remark: Fisher information matrix based observability tool}).

Now we extend Definition \ref{defintion of observablity in maximum likelihood estimation problem} to estimation problems with constraints,
i.e., a noise-free sytem
$\mathbf{Z} = {\mathfrak{F}}(\mathbf{X}) : \mathcal{M}_1^{d_1} \rightarrowtail \mathcal{M}_2^{d_2}$
with constraints
${\mathfrak{G}}(\mathbf{X}) = \mathbf{I} : \mathcal{M}_1^{d_1} \rightarrowtail \mathcal{M}_3^{d_3}$.
If the constraint forms a submanifold $\mathcal{M}_4^{d_4}$ embedded on $\mathcal{M}_1^{d_1}$,
then  we can reduce the constrained MLE to an unconstrained MLE, and verify the observability by Definition \ref{defintion of observablity in maximum likelihood estimation problem}.
The basic tool is the so-called submersion theorem (Proposition 3.3.3 in \cite{absil2009optimization}):
If ${\mathfrak{G}}$ is smooth, and $\mathbf{I}$ is a regular value of ${\mathfrak{G}}$
(i.e. the rank of ${\mathfrak{G}}$ is $d_3$ for each point in $\mathbf{Y} \in \mathcal{M}_1^{d_1}$ satisfying ${\mathfrak{G}} (\mathbf{Y}) = \mathbf{I}$),
then
\begin{equation*}
\mathcal{M}_4 = \{ \mathbf{X}\ \vert 
\ \mathbf{X} \in \mathcal{M}_1^{d_1},
\ {\mathfrak{G}}(\mathbf{X}) = \mathbf{I},
\ \mathbf{rank}({\mathfrak{G}}) = d_3\}
\end{equation*}
admits a differential structure, and $\mathcal{M}_4^{d_4}$ is an embedded submanifold of $\mathcal{M}_1^{d_1}$, with dimension $d_4 = d_1 - d_3$.
If a set of constraints is ``redundant",
we can verify a submanifold by the subimmersion theorem (Proposition 3.3.4 in \cite{absil2009optimization}).

Finally let us examine the case of CB-PGO.
Let $\mathcal{M}_1^{d_1}$ be $\mathrm{SE}(3)^m$.
The $\nu$ constraints clearly satisfy the submersion theorem,
thus the constraints admit a submanifold $\mathcal{M}_4^{d_4}$ embedded in $\mathcal{M}_1^{d_1}$.
The noise-free system, i.e., the measurement function of relative poses
$
\mathbf{Z} =
{\mathfrak{F}}(\mathbf{X}) 
= \mathbf{X}
$,
with
$\mathbf{X} \triangleq \{ \mathbf{T}_{\mathbf{k}} \}_{\mathbf{k} \in \mathcal{E}}$,
is bijective on $\mathcal{M}_1^{d_1}$ thus its restriction to $\mathcal{M}_4^{d_4}$ is also bijective.
Therefore, taking CB-PGO as a MLE on $\mathcal{M}_4^{d_4}$,
we verify CB-PGO to be observable by
Definition \ref{defintion of observablity in maximum likelihood estimation problem}.

\begin{remark}
	\label{remark: Fisher information matrix based observability tool}
	In estimation theory, a common practice is to define the observability of an estimation problem as the invertibility of the FIM \cite{bar2004estimation},
	(also see \cite{andrade2005effects, wang2008observability} for applications in robotics).
	While FIM being a statistical tool, which is related to a specific noise model (like Gaussian),
	the deterministic definition of observability for MLE in Definition \ref{defintion of observablity in maximum likelihood estimation problem} has been shown to be equivalent to FIM based observability if the probability density function has a continuous derivative \cite{jauffret2007observability}.
\end{remark}

\begin{remark}
	The covariance matrix (whose inversion is FIM) of CB-PGO can be computed in closed form (see \cite{gorman1988lower, marzetta1993simple, bai2018robust}).
	However, this covariance matrix is always rank deficient because of the over-parameterization in (\ref{ConstrainedFormulation_PGO}),
	namely $\lvert \mathcal{E} \rvert> \nu$.
	This does not mean CB-PGO is unobservable. To apply FIM based tool correctly,
	we have to obtain FIM for $\mathcal{M}_4^{d_4}$ in its Euclidean space via an atlas \cite{absil2009optimization},
	instead of using FIM on $\mathcal{M}_1^{d_1}$.
	Nevertheless, we can verify the observability by Definition \ref{defintion of observablity in maximum likelihood estimation problem}, without explicitly assigning the atlas.
\end{remark}

\subsection{Jacobian Matrix Design: MCB and Invariance}
\label{section: Jacobian Matrix: Invariance and MCB}

In CB-PGO,
the entry in the Jacobian matrix (Eq. (\ref{eq. cycle-pgo linearied gemometric constraints.}) in Appendix \ref{appendix cycle-pgo Linearization of Metric Cycles}) takes the form of
$
\sigma(\mathbf{k}^c)
\mathbf{Ad}\left( \boldsymbol{\mathcal{P}} (\alpha(\sigma(\mathbf{k}^c))) \right)
$,
with $\boldsymbol{\mathcal{P}} (\cdot)$ being a geometric path inside the cycle.
Let the corresponding topological path be ${\mathcal{P}} (\cdot)$.
Ideally, we want ${\mathcal{P}} (\cdot)$ to be as short as possible, so that less errors will be accumulated in $\boldsymbol{\mathcal{P}} (\cdot)$,
and the Jacobian can more accurately capture the local structure at the linearization point.
By using a MCB, the average length of ${\mathcal{P}} (\cdot)$ is minimized along with the overall length of cycles.
Therefore CB-PGO based on a MCB can be expected to perform better than that based on cycle bases like a FCB.
This is true as will be experimentally validated in
Fig. \ref{figure: global convergence behavior of using different cycle basis algorithm.} and
Fig. \ref{fig: The convergence of each methods on standard benchmarks}.

In CB-PGO, we can minimize the linearization errors inside the Jacobian matrix by using a MCB instead of a FCB.
The idea of having a Jacobian matrix less relevant to linearization errors has been exploited
in the context of Lie group estimation with ``invariance" \cite{barrau:hal-01671724, barrau2018invariant, barrau2015ekf, zhang2017convergence, chauchat2018invariant} as well.
In brief, invariance works by choosing a special Lie group parameterization (i.e., group affine \cite{barrau:hal-01671724}) that
the Jacobian matrix obtained via linearization at a certain point is merely related to some ``error-states" rather than the linearization point directly.
As a result the left/right-hand Jacobian in the BCH formula with respect to the error-states is eliminated.
This technique can significantly improve the convergence of the estimation problem, especially when facing large noise scenarios.

\subsection{Convergence Rate}

	Regarding constrained optimization, one of the major concerns is its convergence rate.
	However, given that
	 CB-PGO in (\ref{ConstrainedFormulation_PGO}) is essentially an equality constrained least squares optimization problem, techniques like SQP can attain quadratic/superlinear convergence \cite{boggs1995sequential}.
	Here we briefly review some work in this line to confirm the claim.
	Experimental validations are provided in Fig. \ref{fig: The convergence of each methods on standard benchmarks}.

For equality constrained least squares optimization,
it has been shown in \cite{tapia1978quasi} that a quadratic convergence rate can be obtained by applying Newton's method (an iterative method to solve a nonlinear equation \cite{ortega1970iterative}) to calculate a critical point of the corresponding Lagrangian function.
This extends the quadratic convergence of Newton's method used in unconstrained optimization to constrained cases by SQP.
Methods based on approximate Lagrangian Hessians can attain superlinear convergence
\cite{boggs1982local, fontecilla1987convergence, thomas1990characterizations}.

In the context of least squares,
the Gauss-Newton method yields quadratic convergence if the residual error at the minimum is zero \cite{madsen1999methods}.
The same conclusion stands in the case of equality constrained least squares, which was proved in \cite{schwetlick1985gauss}.
This convergence result implies that SQP by linearizing the cost function and constraints (like in (\ref{cycle-pgo linearized quadratic programming})) can have basically similar convergence as the Gauss-Newton method.

These justifications are mostly true if the algorithm works around a minimum.
In practice, if the noise level in relative poses is reasonable,
CB-PGO initialized by the measurements of relative poses is close to the ground-truth.
Therefore CB-PGO can have a better convergence (compared with VB-PGO initialized by odometry) because the working point is closer to a minimum and the Hessian matrix is more accurate.

\section{Implementation Details}
\label{tro: section: implemenation details}

The most well-known open-source implementation of a MCB comes from Dimitrios Michail
\cite{mehlhorn2007implementing},
based on the LEDA graph library.
However, the code is not maintained anymore for recent LEDA versions, or Ubuntu systems.
The other competitive implementations, for example \cite{amaldi2010efficient, dutta2017applications}, are not available in open-source.
In contrast, our implementation is freely available to the research community,
and can be easily used in other MCB related problems.

We implement the MCB algorithm described in Section \ref{tro: section: minimum length cycle basis} from scratch with C++ Standard Library and C++ Standard Template Library (STL) only.
For Dijkstra, we use the priority-queue implementation of STL.
We use a dense square matrix with backward pointers to parent vertices to describe the APSP trees.
For set operations on support vectors and cycles, sorted sparse vectors are used instead of binary search trees, which is faster by our experiments.
OpenMP is used to parallelize CPU computations on multiple cores.

Both CB-PGO and VB-PGO are implemented on an open-source graph optimization library, i.e. SLAM++ \cite{ila2017slam++}, which provides the basic operations on block matrices and Cholesky factorization.
In SLAM++, we use the default block Cholesky based linear solver, and the approximate minimum degree ordering algorithm (AMD) \cite{amestoy2003algorithm} to reduce the fill-in.
Both CB-PGO and VB-PGO use the same Lie group implementation,
to avoid the impact of latent numerical round-off errors.

For a fair comparison, we implement chordal initialization
\cite{martinec2007robust, carlone2015initialization}
as an alternative initialization technique, using Block Cholesky factorization.
Instead of initializing the rotational part of poses only \cite{carlone2015initialization}, we reestimate the translational part as well.
This will give an accurate initial objective value for the chordal based methods.
An initialization of relative poses is obtained by recalculating relative poses with pose estimates,
which can be used to initialize CB-PGO.

Noting that for a batch algorithm, we only need to run MCB and AMD once,
which will be followed by several iterations of linearization, Cholesky factorization, and state updates.

\section{Experimental Results}
\label{tro: section: experimental results}

\begin{figure*}[t]
	\centering
	\begin{tabular}{c c}
		\includegraphics[width=0.42\textwidth]{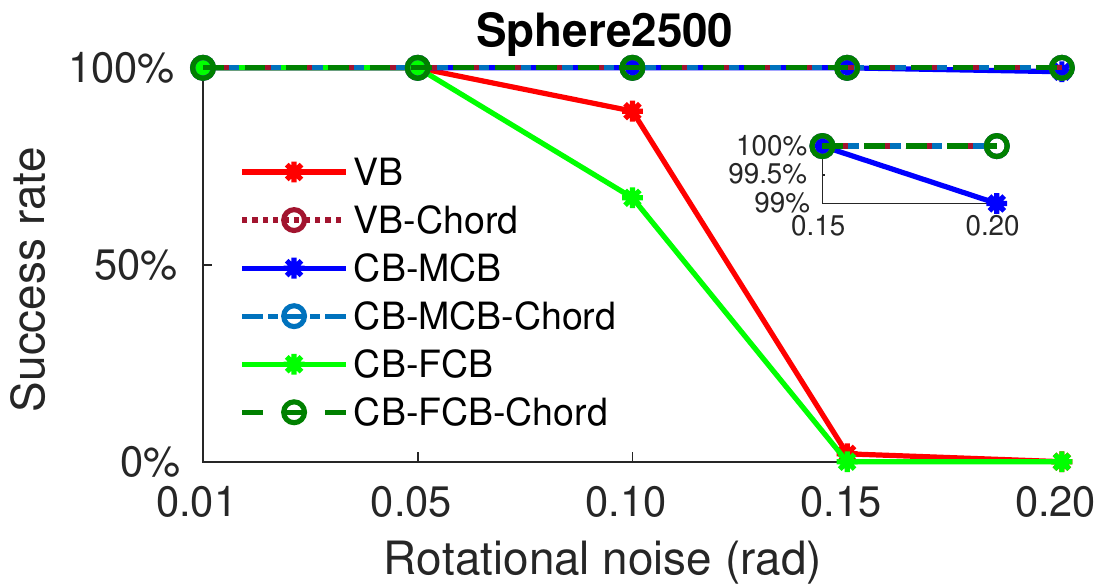}	
		&
		\includegraphics[width=0.42\textwidth]{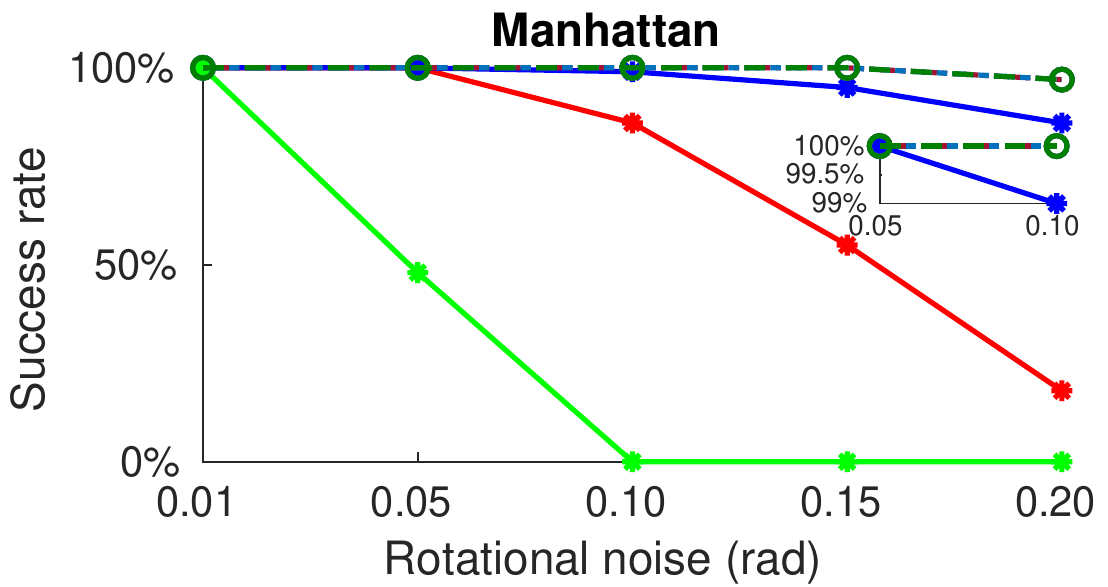}
	\end{tabular}
	\caption{The robustness of vertex-based approaches and cycle based approaches on 100 run Monte-Carlo simulation.
		The curves of the chordal bootstrapped methods, i.e., VB-Chord, CB-MCB-Chord, and CB-FCB-Chord, coincide with one other.}
	\label{figure: global convergence behavior of using different cycle basis algorithm.}
\end{figure*}

\begin{table*}[t]
	\caption{The Computational Time of LexDijkstra and MCB on a Quad-core CPU. Time in Seconds}
	\label{table: LexDijkstra and MCB timing validation on benchmarks.}
	\centering
	\begin{tabular}{@{} ll  cccc  c c c cc @{}}
	\toprule
		&  & \multicolumn{5}{c}{Consistent-APSP} & \multirow{3}{*}{Isometric set} & \multirow{3}{*}{Independence} & \multirow{3}{*}{Proposed} & \multirow{3}{*}{Michail
			\cite{mehlhorn2007implementing}}  \\ 
		\cmidrule{3-7}
		&  & \multicolumn{4}{c}{Method in \cite{hartvigsen1994all}} & \multirow{2}{*}{LexDijkstra} & & & &  \\
		\cmidrule{3-6}
		&  & Dijkstra & Sorting & Processing & Overall &  &  &  &  &  \\
		\midrule
		{MITb} & Sequel & 1.17e-4 & 6.23e-5 & 5.97e-5 &  2.39e-4 & 1.25e-4 & 5.76e-5 & 7.24e-6 &  &           \\
		& Parallel & 6.50e-4        & - & - & 7.72e-4 & 7.20e-4 & 4.61e-5 & - &  8.45e-4 & 2.69e-3   \\	
		\midrule
		{INTEL\_P} & Sequel & 2.42e-3 & 5.06e-4 & 8.66e-4 & 3.79e-3 & 2.79e-3 & 9.01e-4 & 5.35e-5 & &          \\
		& Parallel & 1.78e-3 & - & - & 3.15e-3 & 1.31e-3  & 5.46e-4 & - & 2.11e-3 & 2.46e-2  \\
		\midrule	
		{KITTI} & Sequel & 1.36e-3 & 6.50e-4 & 1.41e-3 & 3.42e-3 & 2.32e-3 & 1.13e-3 & 9.49e-5 & &   \\
		& Parallel & 2.63e-3 & - & - & 4.69e-3 & 2.74e-3    & 7.34e-4 & - & 3.94e-3 & fail  \\
		\midrule
		{Intel} & Sequel & 3.52e-2 & 1.26e-2 & 2.66e-2 & 7.44e-2 & 5.83e-2 & 1.49e-2 & 1.88e-4 & &    \\
		& Parallel & 1.43e-2 & - & - & 5.35e-2 & 2.45e-2    & 8.47e-3 & - & 3.36e-2 & 7.95e-2 \\	\midrule	 	                     	
		{Manhattan} & Sequel & 0.495 & 0.250 & 0.511 & 1.26 & 0.623 & 0.211 & 5.66e-4 &  &           \\
		& Parallel & 0.187 & - & - & 0.948 & 0.241   & 0.105 & - & 0.348 & 0.598 \\
		\midrule
		{Sphere2500} & Sequel & 0.469 & 0.202 & 0.601 & 1.27 & 2.03 & 0.215 & 2.45e-4 & &           \\
		& Parallel & 0.187 & - & - & 0.990 & 0.766   & 0.116 & - &  0.883 & 0.309	 \\	
		\midrule
		{City10k} & Sequel & 7.74 & 3.45 & 11.1 & 22.3 & 11.7 & 4.69 & 6.25e-3 & &  \\
		& Parallel & 2.88 & - & - & 17.4 & 4.53    & 2.71 & - & 7.26 & 8.42  \\
		\midrule
		{Torus10k} & Sequel & 8.91 & 3.64 & 13.6 & 26.2 & 14.3 & 5.93 & 1.08e-2 & &  \\
		& Parallel & 3.36 & - & - & 20.6 & 5.64      &3.29 & - & 8.95 & 14.3   \\	
	\bottomrule
	\end{tabular}
\end{table*}

\begin{figure*}[ht]
	\centering
	\begin{tabular}{cccc}
		\includegraphics[width=.40\textwidth]{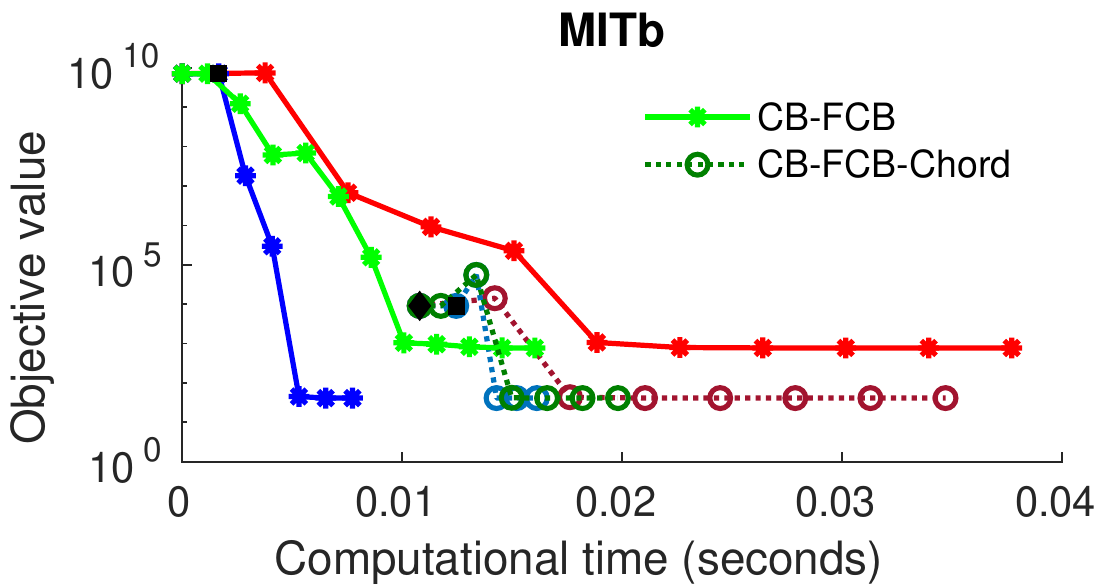}
		&
		\includegraphics[width=.40\textwidth]{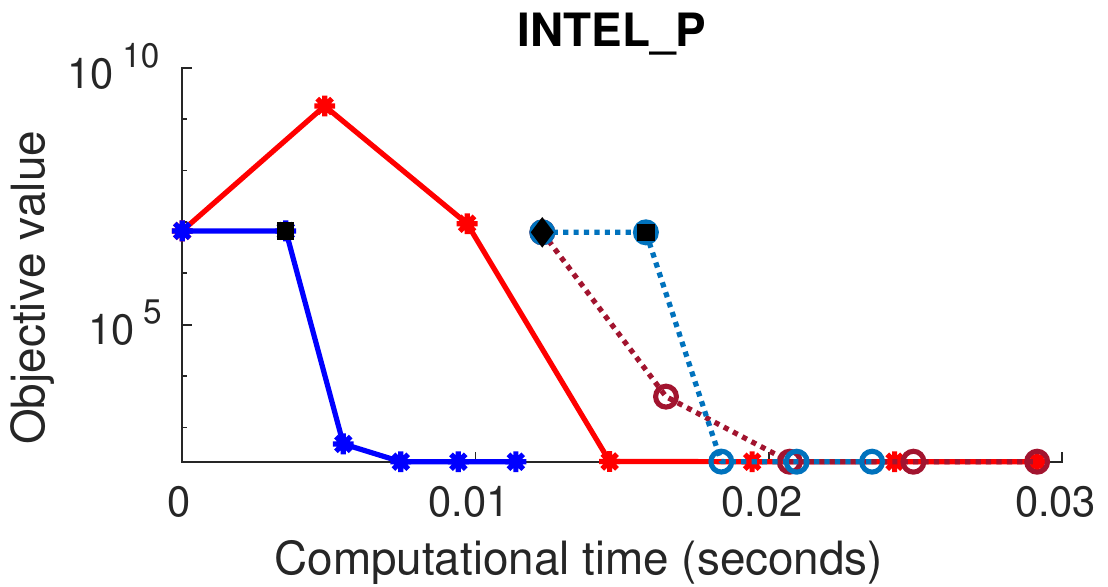} \\
		\includegraphics[width=.40\textwidth]{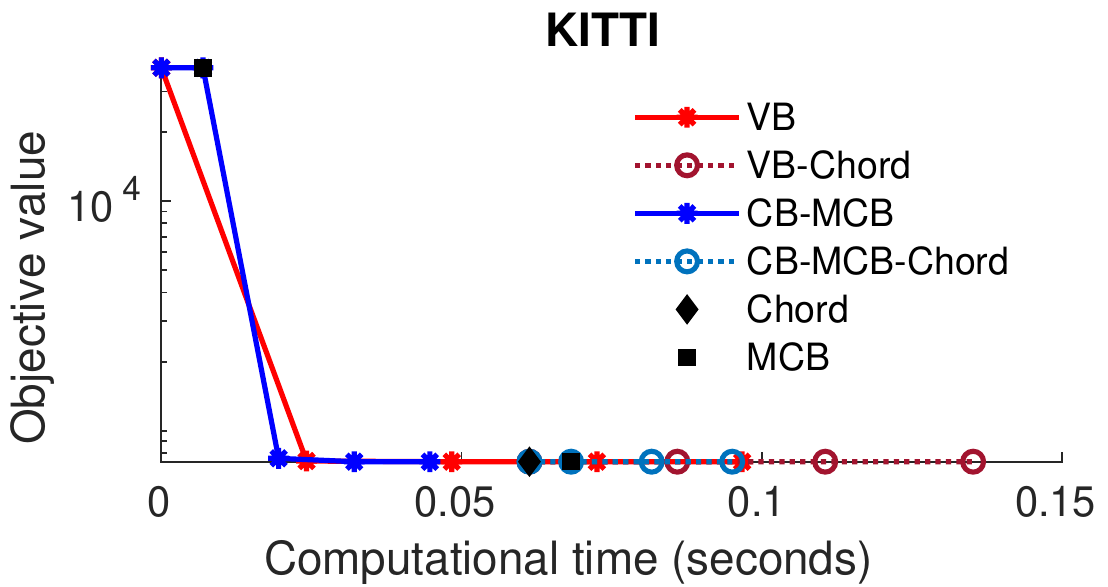}
		& 	
		\includegraphics[width=.40\textwidth]{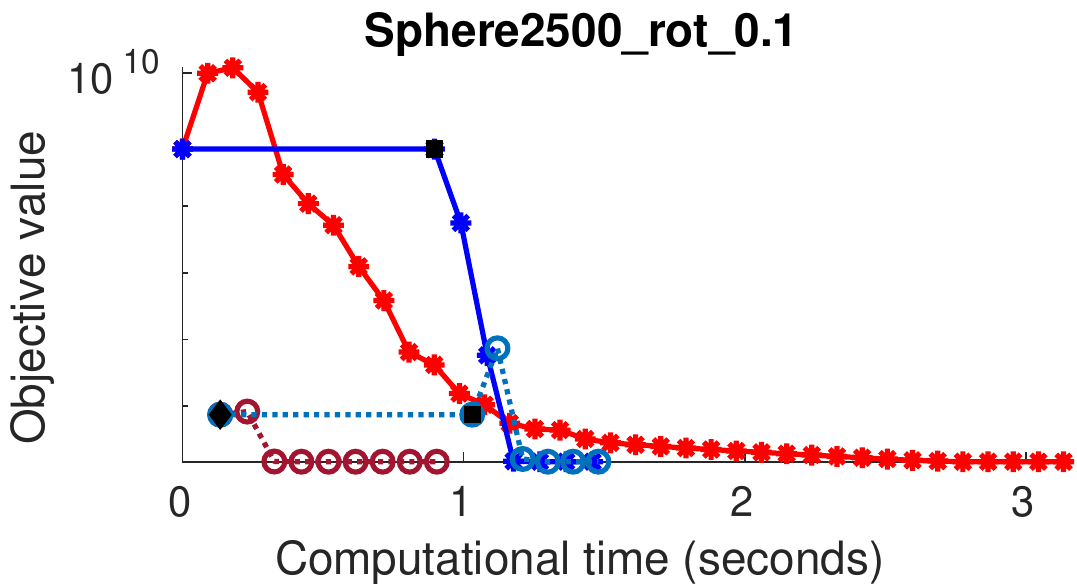}	
	\end{tabular}
	\caption{	
		The convergence of different methods on benchmark datasets.
		Noet that for INTEL\_P, the initial objective values with and without the chordal initialization are	6281560 and 6700310 respectively, which are close but not the same.
		Aside from the MCB,
		the memory usage of VB and CB are comparable.
		For example in INTEL\_P,
		the graph storage takes 1.13MB, while the matrix to be factorized accounts for 8.11MB;
		These numbers in CB are
		1.34MB and 8.04MB respectively.
		For MITb, there are $92$ nonzero blocks in the Cholesky part of CB-MCB, whereas $2462$ for VB.
		For INTEL\_P, CB-MCB accounts for $2234$ nonzero blocks, and VB $4194$.
		The numbers for KITTI are $5015$ and $13831$ respectively.
		For dense graphs like Sphere2500, there are $12244$ nonzero blocks in the Cholesky part of CB-MCB, while the number for VB is $12398$.
	}
	\label{fig: The convergence of each methods on standard benchmarks}
\end{figure*}

\begin{table*}[ht]
	\caption{Computational Time of Chordal Initialization, MCB, and Cholesky Factorization on Benchmark Datasets.
	$\mathcal{G}(\mathcal{V}, \mathcal{E})$ stands for the original graph, while $ \bar{\mathcal{G}}(\bar{\mathcal{V}}, \bar{\mathcal{E}})$ the reduced weighted graph.
	Time in Seconds}
	\label{table: comparision of timing of ECPGO and NEPGO on standard benchmarks}
	\centering
	\begin{tabular}{@{\extracolsep{4pt}} l  cccc   c  c    c    c    c c c  @{}}
	\toprule
		& \multicolumn{4}{c}{${\mathcal{G}}$} & \multicolumn{2}{c}{$\bar{\mathcal{G}}$}  & Chord. Init. & MCB & \multicolumn{3}{c}{Cholesky Per Iter.}   \\
		\cmidrule{2-5} 
		\cmidrule{6-7}
		\cmidrule{10-12}
		&  $\lvert \mathcal{E} \rvert$   &  $\lvert \mathcal{V} \rvert$  & $\nu$ & $\nu/\lvert \mathcal{E} \rvert$ &  $\lvert \bar{\mathcal{E}} \rvert$  &  $\lvert \bar{\mathcal{V}} \rvert$   &	 all    & CB-MCB    & CB-MCB & CB-FCB & VB 	\\
		\midrule
		\textbf{MITb}    &  827  &  808  & 20  & \textbf{2.42\%}  &  60   &  41   & 7.32e-3  & 8.45e-4 & 6.74e-5 & 6.12e-5 & 1.31e-3  \\
		\textbf{INTEL\_P}& 1483  &  1228 &  256&  \textbf{17.3\%} &  396  & 141 & 1.18e-2
		& 2.11e-3 &  1.04e-3 & 3.98e-2 &  2.28e-3 \\		
		\textbf{KITTI}   & 5065  & 4541  & 525 &   \textbf{10.4\%} &  654  &  130  & 5.98e-2  & 3.94e-3 &  2.60e-3 & 1.43 & 1.02e-2  \\
		Intel            & 1835  &  943 & 893 &  48.7\% &  1515   & 623 & 1.45e-2 & 3.36e-2 &  3.31e-3 & 0.985 &  2.75e-3  \\
		\midrule
		Manhattan      	 & 5453  &  3500 &1954 &  35.8\% &  4350  & 2397 &  4.43e-2 & 0.348 & 8.16e-3 & 1.31 & 8.42e-3 \\
		Sphere2500 	     & 4949  &  2500 &2450 &  49.5\% &   4947  & 2498 & 0.131 & 0.883 & 8.51e-2 &  0.262 & 8.20e-2    \\
		City10k    	     & 20687 & 10000 &10688&  51.7\% & 19528 & 8841 & 0.209 & 7.26 & 6.78e-2 & fail & 6.11e-2	\\
		Torus10k 	     & 22280 & 10000 &12281&  55.1\% &  21542 & 9262 & 0.589 & 8.95 & 0.319 & fail & 0.340  \\ 
	\bottomrule
	\end{tabular}
\end{table*}

In this paper,
the experiments are carried out by a laptop equipped with a quad-core CPU,
Intel(R) Core(TM) i5-5300U CPU @ 2.30GHz $\times$ 4,
running on Ubuntu 16.04 LTS.

We will use four real datasets and four simulated datasets as standard benchmarks.
INTEL\_P, Intel and MITb are
obtained by processing the raw measurements from wheel odometry and laser range finder \cite{carlone2014angular}.
Raw data are available at \cite{carlone2011rawdata}.
KITTI is a pose graph generated by the proSLAM framework \cite{schlegel2018proslam} from the vision benchmark dataset \cite{geiger2012we}.
The four simulated datasets and their creators are: Manhattan \cite{olson2006fast}, City10k \cite{kaess2008isam},
Sphere2500 \cite{kaess2008isam}, and Torus10000 \cite{kaess2008isam}.

We will use the \textsf{cycle ratio} of a graph, defined as $\nu / \lvert \mathcal{E} \rvert$,
to benchmark the graph sparsity.
In graph simulations, we regard the local minimum computed by using the ground-truth initialization as the \textsf{global minimum}.
Let the objective value at the global minimum be $f^{\star}$.
For any computed solution, if its objective value $f$ satisfies $\lvert f / f^{\star} - 1 \rvert < 0.01$, the solution is counted as a \textsf{success}.
The number of successes divided by the number of Monte Carlo runs is defined as the \textsf{success rate}.
For CB-PGO, at each iteration, we firstly calculate a pose configuration through odometry
using the current estimate of relative-poses.
Then we substitute the pose configuration to the cost function of VB-PGO, and use the obtained cost as the \textsf{objective value of CB-PGO}.

In this section,
the VB-PGO initialized by odometry is denoted as VB, while CB-PGO initialized by the measurements of relative poses is denoted as CB.
We distinguish CB-PGO techniques based on the MCB and FCB by denoting them as CB-MCB, CB-FCB respectively.
Besides the ``natural" initialization for VB-PGO and CB-PGO,
we use chordal initialization to bootstrap the rotational part \cite{carlone2015initialization}.
To evaluate the objective value after chordal initialization,
we recompute the optimal translational estimate after the rotational initialization.
The chordal bootstrapped VB, and CB variants are termed as VB-Chord, CB-MCB-Chord and CB-FCB-Chord respectively.

The maximal iterations are set to 50, and we stop the algorithm if the perturbation (i.e., state updates) has a norm less than $0.001$.
For CB-PGO, we also ensure the constraint residual norm to be less than $0.001$.

%
%
%

\subsection{MCB}

The timing statistics of the proposed MCB is reported in Table \ref{table: LexDijkstra and MCB timing validation on benchmarks.}.
The proposed MCB algorithm consists of three major parts: consistent-APSP, isometric set construction (Section \ref{subsection: superset of MCB: isometric circuits}), and independence test (Section \ref{sec: mcb: independecne test}).
The construction of
Horton set (Section \ref{section: mcb: Superset of MCB: Horton Set})
requires a APSP, while its reduction to the isometric set requires a consistent-APSP.
In Table \ref{table: LexDijkstra and MCB timing validation on benchmarks.},
Dijkstra is used to compute a APSP. The consistent-APSP can be computed via the method in \cite{hartvigsen1994all}, or by LexDijkstra (proposed).
All the timings regarding these two consistent-APSP algorithms are presented.
To benchmark the overall effectiveness of the proposed MCB, we compare with the state-of-the-art open-source implementation in \cite{mehlhorn2007implementing}.
In Table \ref{table: LexDijkstra and MCB timing validation on benchmarks.}, the symbol ``-" in the ``parallel" row means that the corresponding computation cannot be parallelized.

In Table \ref{table: LexDijkstra and MCB timing validation on benchmarks.},
the proposed MCB algorithm is at least as fast as the state-of-the-art in \cite{mehlhorn2007implementing}.
Actually the proposed MCB is much faster for sparse graphs (due to the pruning of degree-two vertices), and slightly faster for most dense graphs, except for the Sphere dataset.
This is due to the high-symmetry of the graph which results to many lexicographical comparisons.
The implementation in \cite{mehlhorn2007implementing} fails at the KITTI dataset, because it cannot handle parallel edges.
In addition,
Table \ref{table: LexDijkstra and MCB timing validation on benchmarks.}
 clearly shows the advantage of using the proposed LexDijkstra to compute a consistent-APSP, since it avoids the sequel bottleneck in \cite{hartvigsen1994all}.
In the worst case, LexDijkstra accounts for $1.5$ times the timing of the pure Dijkstra (APSP),
which we believe is already close to the lower borderline.
Note that this timing will double without Theorem \ref{theorem: LexDijkstra, only need to traverse back to a common node.}.

In Table \ref{table: LexDijkstra and MCB timing validation on benchmarks.},
the computation is parallelized by a quad-core CPU; however these timings can be more advantageous if the CPU has more cores (like an eight-core CPU or more).

\subsection{Standard PGO Benchmarks}

\subsubsection{Statistics of Each Part}
\label{experimental results, standard benchmark, section timing}

We report the computational time for Cholesky factorization, MCB, and Chordal initialization in Table \ref{table: comparision of timing of ECPGO and NEPGO on standard benchmarks}.
The timings for linearization, state updates, and system matrix constructions are ignored,
because these operations are rather cheap compared with the Cholesky part.
An exception is CB-PGO using FCB (CB-FCB):
The system matrix allocation in this case can be rather expensive by indexing non-zero elements in Jacobian.
In case of
Manhattan, it takes almost $1$ minute to construct the system matrix, and fails in case of City10k and Torus10k.
The size and sparsity of the benchmarks are included as well.

Regarding Cholesky factorization, CB-MCB and VB have comparable performance, while CB-MCB is (2-3 times) faster on sparse graphs.
The Cholesky part of CB-FCB takes around two magnitudes of time compared with that of CB-MCB/VB,
except for the MITb dataset (which has a rather small $\nu=20$ thus it actually does not matter whether the system matrix is sparse or not).
The timing statistics of the Cholesky part indicates that CB-FCB is not a viable approach (2 magnitudes slower than VB in almost any cases).

In Table \ref{table: comparision of timing of ECPGO and NEPGO on standard benchmarks},
the timing of the MCB and Cholesky parts are comparable for sparse graphs,
i.e., the four real dataset,
MITb, INTEL\_P, KITTI, Intel.
Given that we only need to run MCB once, followed by multiple Cholesky iterations,
CB-MCB can take advantage in this case, in particular for the MITb, INTEL\_P and KITTI datasets.
For dense (simulated) graphs, like Sphere, Manhattan, City10k and Torus10k, since the MCB part is too expensive compared with the Cholesky part, it is not a good choice to use CB-MCB (or CB-MCB-Chord) if timing is a critical consideration.

The chordal initialization accounts for roughly 2-3 iterations of the Cholesky time.
This is understandable since its solves a linear system of $k$ times larger (with $k=2$ for 2D; and $k=3$ for 3D).
We will use chordal initialization to boost all the methods (termed VB-Chord, CB-MCB-Chord, CB-FCB-Chord) to examine the robustness.

\begin{figure*}[t]
	\centering
	\begin{tabular}{cccc}
		\includegraphics[width=0.40\textwidth]{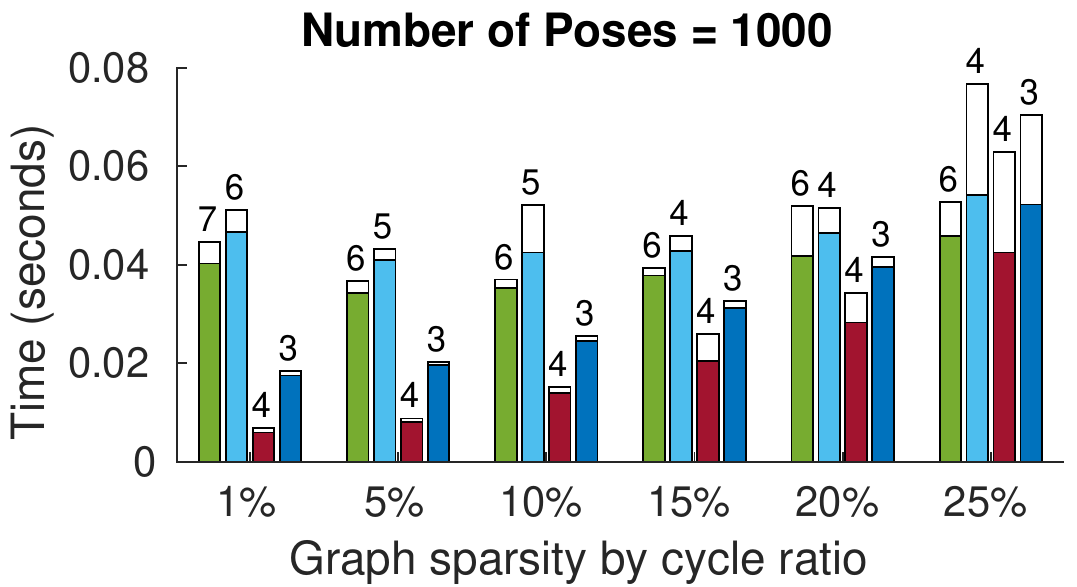} &
		 \includegraphics[width=0.40\textwidth]{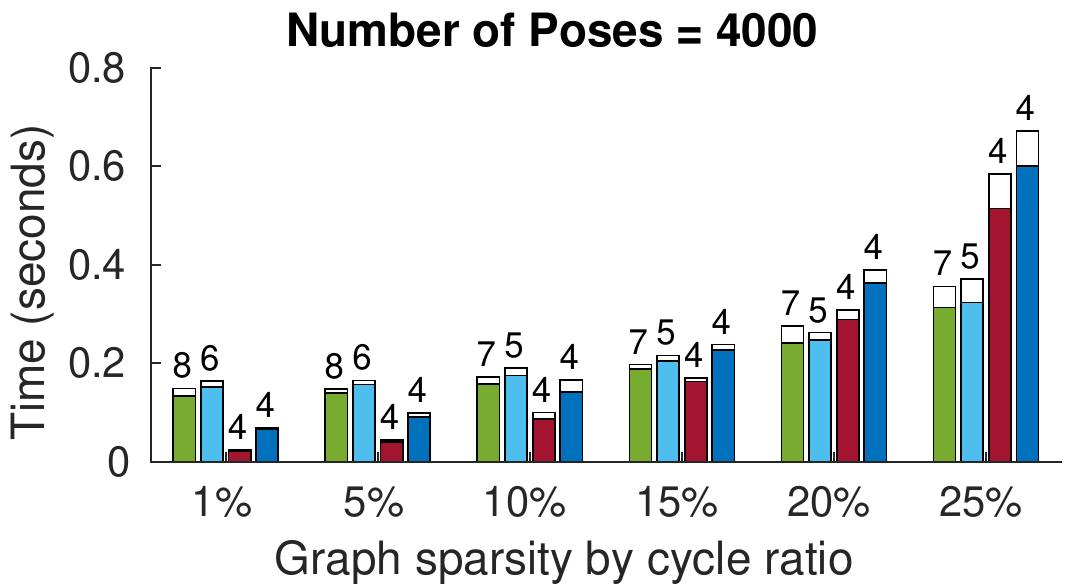} \\
		\includegraphics[width=0.40\textwidth]{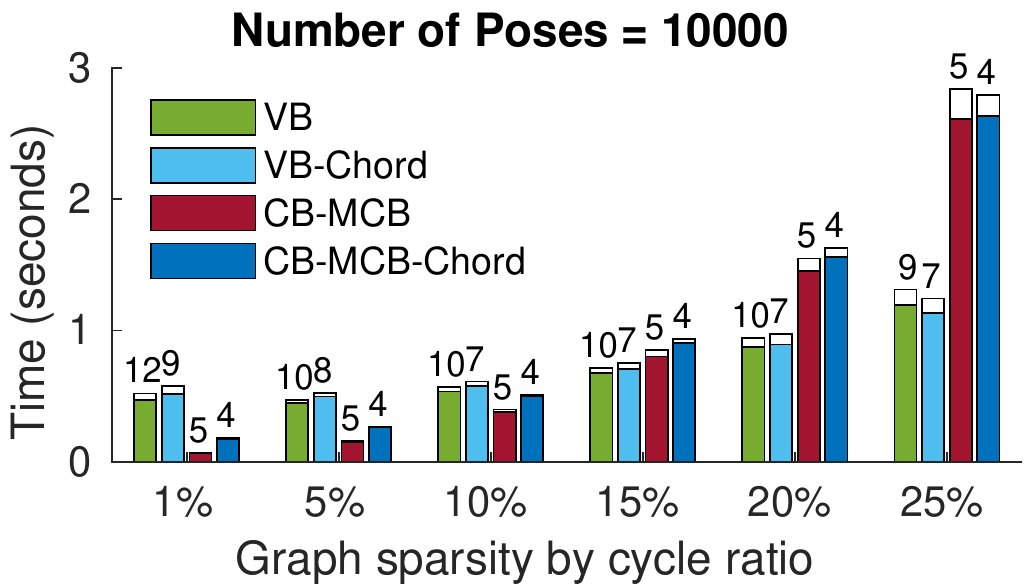} &
		 \includegraphics[width=0.40\textwidth]{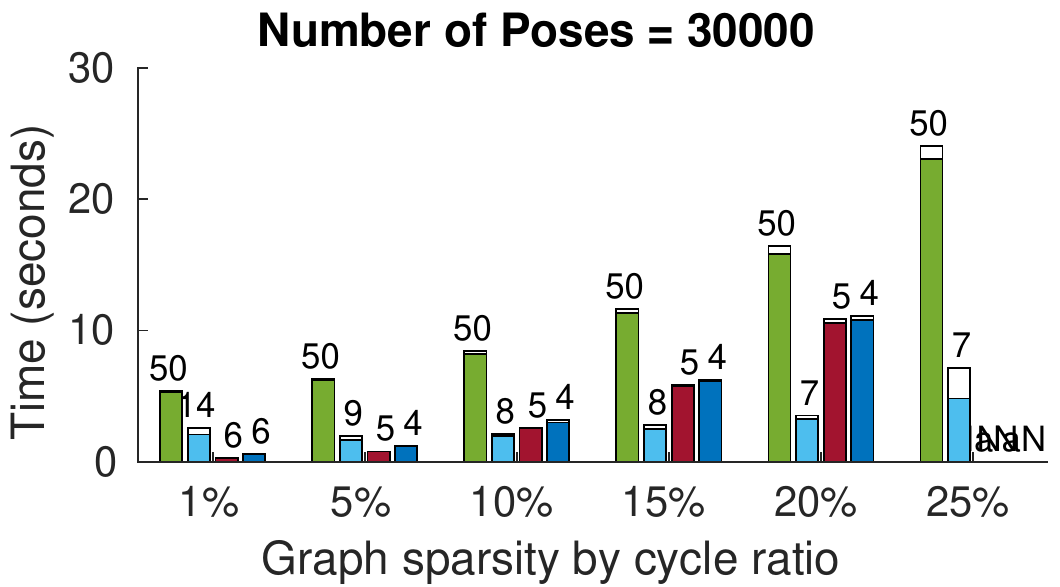}
	\end{tabular}
	\caption{The computational time of VB (VB-PGO initialized by odometry),
		VB-Chord (VB-PGO initialized by the Chordal initialization technique),
		CB-MCB (CB-PGO based on the MCB, initialized by the measurements of relative poses),
		and CB-MCB-Chord (CB-PGO based on the MCB, initialized by the Chordal initialization technique)
		on 100 run Monte-Carlo simulation.
		The mean is plotted with color, and the standard deviation is added to the bar-plot as the white margin.
		The number above each bar is the used iterations.
		The maximal iteration is set to $50$.
	}
	\label{figure: comparision of timing of ECPGO and NEPGO on simulation datasets}	
\end{figure*}

\subsubsection{Convergence and Complexity}

	 In Fig.
	 \ref{fig: The convergence of each methods on standard benchmarks},
	we visualize the convergence of VB, VB-Chord, CB-MCB, CB-MCB-Chord, CB-FCB, and CB-FCB-Chord, against their computational time.
	The time point of the MCB and Chordal initialization are marked out for a clear visualization.

	The MTIb dataset shows a case where VB and CB-FCB converges to a local minimum,
	where CB-MCB converges to the global minimum.
	The chordal bootstrapped approaches, VB-Chord and CB-FCB-Chord, can converge to the global minimum as well, while taking a longer time.
	In this case, CB-MCB is both the fastest and robustest approach.
	The fast convergence of CB-MCB is further validated by a sparse graph, INTEL\_P ($17.3\%$ sparsity),
	where it significantly outperforms VB, and VB-Chord.
	The results for FCB, FCB-Chord are not shown for INTEL\_P as the the timings are too large to be plotted on the same scale.
	Similar results are witnessed on the KITTI dataset, while the convergence is much faster given that the dataset is less noisy.
	For dense graphs, CB-MCB is not advantageous in terms of the computational time, since the MCB part is dominant over the Cholesky part as shown in Table
	\ref{table: comparision of timing of ECPGO and NEPGO on standard benchmarks}.
	However, even in these cases, CB-MCB is still useful because it yields a faster convergence.
	For instance on the Sphere2500 dataset,
	if we increase the noise level to 0.1 rads in the rotation part, the convergence of VB degenerates dramatically, and CB-MCB can outperform VB because of the faster convergence.
	Nonetheless, the VB-Chord can significantly improve the convergence of VB, thus yielding a smaller timing.

\subsection{Monte-Carlo Simulation}

\subsubsection{Global Minimum}

To examine the robustness of CB-PGO, in terms of converging to the global minimum,
we provide a Monte-Carlo simulation
on the simulated dataset Manhattan (2D), and Sphere2500 (3D).

	We recreate the dataset from its ground-truth with additive noise in the exponential coordinate,
	with $0.1\mathsf{m}$ standard deviation on the translational part,
	and
	$0.01\mathsf{rads}$
	$0.05\mathsf{rads}$,
	$0.10\mathsf{rads}$,
	$0.15\mathsf{rads}$,
	$0.20\mathsf{rads}$
	respectively on the rotational part.

	For each case, we generate $100$ noisy graphs,
	and report in Fig. \ref{figure: global convergence behavior of using different cycle basis algorithm.} the success rate of VB, VB-Chord, CB-MCB, CB-MCB-Chord, CB-FCB, and CB-FCB-Chord,
	when solving these graphs.
	From Fig. \ref{figure: global convergence behavior of using different cycle basis algorithm.},
	we conclude that
	CB-FCB is basically worse than VB, while CB-MCB is much more robust than VB (initialized by odometry) and CB-FCB.
	Unsurprisingly, the chordal bootstrapped approaches, i.e., VB-Chord, CB-MCB-Chord and CB-FCB-Chord show the best robustness.
	However, a pure CB-MCB (initialized by relative measurements), is almost as robust as the chordal bootstrapped approaches.

\subsubsection{Computational Time}

To better understand the computational complexity of VB, VB-Chord, CB-MCB and CB-MCB-Chord on sparse graphs,
we perform a Monte-Carlo simulation,
with respect to different cycle ratios
and different number of poses.
Cases for CB-FCB, CB-FCB-Chord are ignored because the FCB based approaches do not scale well with respect to graph topologies (see Table \ref{table: comparision of timing of ECPGO and NEPGO on standard benchmarks}).

We firstly simulate a giant dense graph using the g2o simulator \cite{kummerle2011g} (g2o\_simulator3d).
Then the graph is tailored respectively to $1000$, $4000$, $10000$ and $30000$ poses,
and pruned to different sparsity, i.e., $1\%$, $5\%$, $10\%$, $15\%$, $20\%$ and $25\%$.
The process randomly generates 100 graphs for each pose number and cycle ratio combination.

The running time statistics of VB, VB-Chord, CB-MCB and CB-MCB-Chord for each simulated scenario is recorded in
Fig. \ref{figure: comparision of timing of ECPGO and NEPGO on simulation datasets},
where the mean is plotted with color, and the standard deviation is added to the bar-plot as the white margin.
The used iterations are also included above each bar.
Note that the maximal iteration is set to $50$.

In Fig. \ref{figure: comparision of timing of ECPGO and NEPGO on simulation datasets},
CB-MCB is obviously more advantageous than VB for all the tested scenarios with a maximal cycle ratio at around $15\%$, as well as VB-Chord at round $10\%$.
When the graph becomes denser, with the cycle ratio exceeding $20\%$, the timings of CB-MCB and CB-MCB-Chord grow with respect to the number of poses as their MCB parts start to dominate the complexity.
However, even in these cases, the timings of CB-MCB and CB-MCB-Chord are still comparable to that of VB and VB-Chord, without a dramatical deterioration.
The VB-Chord can improve the convergence of VB with a slight trade-off on the computational cost.
The same applies to CB-MCB and CB-MCB-Chord, while the improvement is less obvious.
Considering CB-MCB is almost as robust as chordal based approaches, as shown in Fig. \ref{figure: global convergence behavior of using different cycle basis algorithm.},
it seems that CB-MCB-Chord does not offer much compared to CB-MCB.

In case of huge graphs, like $30000$ poses,
the numerical stability of VB degenerates dramatically, as can be clearly seen by the iterations consumed.
In contrast, the CB-MCB shows a much better numerically stability and convergence property in this scenario,
while the bootstrapped approach VB-Chord can significantly improve the convergence of VB.
It is worth noting that CB-MCB and CB-MCB-Chord can fail when allocating the memory for shortest path trees (a dense matrix of $O(n^2)$ memory) if the reduced graph still has too many vertices.
Such a case is shown at $25\%$ sparsity, with 30000 poses.

\section{Conclusion}
\label{tro: section: conlcusion}

To summarize, we propose CB-PGO, a robust and efficient PGO technique that works in the cycle space of the graph.
We characterize the graph sparsity by a MCB, which reduces the numerical complexity and enhances the convergence to the global minimum.
We design a tailored MCB algorithm for sparse, positive-integer weighted graphs, which can be used for other cycle based applications as well.
The claims on the convergence and computational complexity are validated with experiments.
We provide an open-source C++ implementation that is freely available to the community to benefit future research in this direction.
The future work includes:
extending the cycle-based approach to other sparse graph instances;
exploiting sparse representations for the consistent APSP storage;
developing incremental algorithms for cycle based approaches;
exploring the possibility of using
Frobenius norm based cost function and convex relaxations.

\begin{appendices}

	\section{Linearization of Cycle Based PGO}
	\label{appendix cycle-pgo Linearization of Cycle Based PGO}

	In section, we provide the computational details in terms of how to linearize the CB-PGO formulation in (\ref{ConstrainedFormulation_PGO}).

	\subsection{Linearization of Cost Function}
	\label{appendix cycle-pgo Linearization of Cost Function}
	Denote the error of each edge $\mathbf{k}$ at its estimate $\hat{\mathbf{T}}_{\mathbf{k}}$ to be
	$\boldsymbol{\eta}_{\mathbf{k}} = \mathbf{Log}
	(
	\mathbf{\tilde{T}}_{\mathbf{k}}^{-1} \cdot \hat{\mathbf{T}}_{\mathbf{k}} 
	)
	$.
	The linearization of the cost function is trivial using the approximate BCH formula:
	\begin{equation*}
	\begin{aligned}
	& \sum_{ \mathbf{k} \in \mathcal{E} } 
	\lVert   
	\mathbf{Log}
	(
	\mathbf{\tilde{T}}_{\mathbf{k}}^{-1} \cdot \mathbf{T}_{\mathbf{k}}
	)
	\rVert_{\boldsymbol{\Sigma}_{\mathbf{k}}}^2
	\\ \gets & 
	\sum_{ \mathbf{k} \in \mathcal{E} } 
	\lVert   
	\mathbf{Log}
	\left(
	\mathbf{\tilde{T}}_{\mathbf{k}}^{-1} \cdot \hat{\mathbf{T}}_{\mathbf{k}} \cdot \mathbf{Exp} \left(\boldsymbol{\xi}_{\mathbf{k}}\right)
	\right)
	\rVert_{\boldsymbol{\Sigma}_{\mathbf{k}}}^2
	\\ = & 
	\sum_{ \mathbf{k} \in \mathcal{E} } 
	\lVert   
	\mathbf{Log}
	\left(
	\mathbf{Exp}( \boldsymbol{\eta}_{\mathbf{k}} ) \cdot \mathbf{Exp} ( \boldsymbol{\xi}_{\mathbf{k}} )
	\right)
	\rVert_{\boldsymbol{\Sigma}_{\mathbf{k}}}^2
	\\ \approx & 
	\sum_{ \mathbf{k} \in \mathcal{E} } 
	\lVert   
	\mathbf{Log}
	\left(
	\mathbf{Exp}\left( \boldsymbol{\eta}_{\mathbf{k}}   +  \mathbf{J}_{\mathbf{r}}^{-1} ( \boldsymbol{\eta}_{\mathbf{k}} ) \boldsymbol{\xi}_{\mathbf{k}}\right)
	\right)
	\rVert_{\boldsymbol{\Sigma}_{\mathbf{k}}}^2
	\\ = & 
	\sum_{ \mathbf{k} \in \mathcal{E} } 
	\lVert   
	\boldsymbol{\eta}_{\mathbf{k}}   +  \mathbf{J}_{\mathbf{r}}^{-1} ( \boldsymbol{\eta}_{\mathbf{k}} ) \boldsymbol{\xi}_{\mathbf{k}}
	\rVert_{\boldsymbol{\Sigma}_{\mathbf{k}}}^2
	= 
	\lVert   
	\boldsymbol{\eta}   +  \mathbf{J}^{-1} \boldsymbol{\xi}
	\rVert_{\boldsymbol{\Sigma}}^2
	\end{aligned}
	\end{equation*}
	where, $\boldsymbol{\eta} = \mathbf{stack}\{ \boldsymbol{\eta}_{\mathbf{k}} \}_{ \mathbf{k} \in \mathcal{E} }$,
	$\boldsymbol{\xi} = \mathbf{stack}\{ \boldsymbol{\xi}_{\mathbf{k}} \}_{ \mathbf{k} \in \mathcal{E} }$,
	$\mathbf{J} = \mathbf{blkdiag}\{
	\mathbf{J}_{\mathbf{r}} \left( \boldsymbol{\eta}_{\mathbf{k}} \right)
	\}_{ \mathbf{k} \in \mathcal{E} }
	$,
	and
	$\boldsymbol{\Sigma} = \mathbf{blkdiag}\{ \boldsymbol{\Sigma}_{\mathbf{k}} \}_{ \mathbf{k} \in \mathcal{E} }
	$.

	\subsection{Linearization of Geometric Cycles}
	\label{appendix cycle-pgo Linearization of Metric Cycles}

	Let us take the metric cycle ${\boldsymbol{\mathcal{C}}}^{\mathrm{lhs}} = \mathbf{I}$ as an example.
	Recall that
	$
		{\boldsymbol{\mathcal{C}}}^{\mathrm{lhs}}
		=
		\mathbf{T}_{\mathbf{1}^c}^{\sigma(\mathbf{1}^c)} 
		\mathbf{T}_{\mathbf{2}^c}^{\sigma(\mathbf{2}^c)} 
		\cdots 
		\mathbf{T}_{\boldsymbol{\lambda}^c}^{\sigma(\boldsymbol{\lambda}^c)}
	$,
	where $\mathbf{T}_{\mathbf{1}^c},\dots \mathbf{T}_{\boldsymbol{\lambda}^c}$ are the sequential relative poses obtain by a traversal.
	$\sigma(\mathbf{1}^c),\dots \sigma(\boldsymbol{\lambda}^c)$
	are the relative edge orientations with respect to the traversal, assigned to $+1$ if the traversal uses the edge in the forward direction, and $-1$ otherwise. 

	The constraint ${\boldsymbol{\mathcal{C}}}^{\mathrm{lhs}} = \mathbf{I}$ can be written as
	$ \mathbf{Log}( {\boldsymbol{\mathcal{C}}}^{\mathrm{lhs}} ) = \mathbf{0}$.
	In the vector space of $\mathfrak{se}(3)$,
	the first-order Taylor expansion takes the form,
	\begin{equation*}
	\sum_{ \mathbf{k}^c = \mathbf{1}^c }^{\boldsymbol{\lambda}^c} 
	\frac{\partial  \mathbf{Log}( {\boldsymbol{\mathcal{C}}}^{\mathrm{lhs}} ) }{\partial \boldsymbol{\xi}_{\mathbf{k}^c}^{T} }
	\cdot \boldsymbol{\xi}_{\mathbf{k}^c}
	+
	\mathbf{Log}
	(
	{\boldsymbol{\mathcal{C}}}^{\mathrm{lhs}}
	)
	= \mathbf{0}
	.
	\end{equation*}
	At a set of given estimates $\hat{\mathbf{T}}_{\mathbf{k}^c}$
	$( \mathbf{k}^c = \mathbf{1}^c \cdots \boldsymbol{\lambda}^c )$,
	let us define the error of the geometric cycle to be
	$
	\boldsymbol{\beta}=
	\mathbf{Log}
	(
	{\boldsymbol{\mathcal{C}}}^{\mathrm{lhs}}
	)
	$.
	For any $\mathbf{h}^c \in [\mathbf{1}^c, \boldsymbol{\lambda}^c]$,
	with a bit abuse of notation, we split the geometric cycle into two geometric paths
	$\boldsymbol{\mathcal{P}} (\mathbf{h}^c)$,
	$\boldsymbol{\mathcal{\bar{P}}} (\mathbf{h}^c)$:
	\begin{equation*}
	\boldsymbol{\mathcal{P}} (\mathbf{h}^c) =
			\hat{\mathbf{T}}_{\mathbf{1}^c}^{\sigma(\mathbf{1}^c)} 
			\cdots 
			\hat{\mathbf{T}}_{\mathbf{h}^c}^{\sigma(\mathbf{h}^c)}
			,
	\end{equation*}
	\begin{equation*}
	\boldsymbol{\mathcal{\bar{P}}} (\mathbf{h}^c) = 
				\hat{\mathbf{T}}_{\mathbf{(h+1)}^c}^{\sigma((\mathbf{h+1})^c)} 
				\cdots 
				\hat{\mathbf{T}}_{\boldsymbol{\lambda}^c}^{\sigma(\boldsymbol{\lambda}^c)}
				.
	\end{equation*}
	Let $\boldsymbol{\mathcal{P}} (\mathbf{0}^c) = \mathbf{I}$ be a special case.
	The equality below is trivial
	\begin{equation}
	\label{Trivial Relation of P, Pbar, b}
	\boldsymbol{\mathcal{P}} (\mathbf{h}^c) \cdot \boldsymbol{\mathcal{\bar{P}}} (\mathbf{h}^c)
	= \mathbf{Exp}( \boldsymbol{\beta} )
	.
	\end{equation}

	Now let's add a perturbation $\boldsymbol{\xi}_{\mathbf{k}^c}$ to the edge $\mathbf{k}^c$ inside ${\boldsymbol{\mathcal{C}}}^{\mathrm{lhs}}$ via the exponential mapping,
	\begin{equation*}
	{\boldsymbol{\mathcal{C}}}^{\mathrm{lhs}}
	\gets
	\hat{\mathbf{T}}_{\mathbf{1}^c}^{\sigma(\mathbf{1}^c)} 
	\cdots
	\left( \hat{\mathbf{T}}_{\mathbf{k}^c}   \cdot \mathbf{Exp} (\boldsymbol{\xi}_{\mathbf{k}^c})   \right)^{\sigma(\mathbf{k}^c)} 
	\cdots
	\hat{\mathbf{T}}_{\boldsymbol{\lambda}^c}^{\sigma(\boldsymbol{\lambda}^c)} 
	.
	\end{equation*}
	Let the perturbed ${\boldsymbol{\mathcal{C}}}^{\mathrm{lhs}}$ at $\mathbf{k}^c$ be ${\boldsymbol{\mathcal{C}}}^{\mathrm{lhs}} \vert_{\mathbf{k}^c}$,
	which can be compactly written as,
	\begin{equation*}
	{\boldsymbol{\mathcal{C}}}^{\mathrm{lhs}} \vert_{\mathbf{k}^c}
	=
	\boldsymbol{\mathcal{P}} (\alpha(\sigma(\mathbf{k}^c)))
	\mathbf{Exp} \left( \sigma(\mathbf{k}^c)  \boldsymbol{\xi}_{\mathbf{k}^c}\right) 
	\boldsymbol{\mathcal{\bar{P}}} (\alpha(\sigma(\mathbf{k}^c))) 
	\end{equation*}
	where $\alpha(\sigma(\mathbf{k}^c))$ is a scalar function with respect to $\sigma(\mathbf{k}^c)$,
	\begin{equation*}
	\alpha(\sigma(\mathbf{k}^c)) = 
	\begin{cases}
	\mathbf{k}^c  	 &  \mathrm{if}\ \sigma(\mathbf{k}^c) = +1 \\
	(\mathbf{k-1})^c  &  \mathrm{if}\ \sigma(\mathbf{k}^c) = -1 
	\end{cases}
	.
	\end{equation*}
	
	Applying (\ref{Adjoint Property of Lie Group}) and considering (\ref{Trivial Relation of P, Pbar, b}),
	${\boldsymbol{\mathcal{C}}}^{\mathrm{lhs}} \vert_{\mathbf{k}^c}$ can be written as the multiplication of two matrix exponentials, which can be concatenated by the approximate BCH formula, as
	\begin{equation*}
	\begin{aligned}
	& {\boldsymbol{\mathcal{C}}}^{\mathrm{lhs}} \vert_{\mathbf{k}^c}
	=
	\mathbf{Exp} \left( 
	\sigma(\mathbf{k}^c)
	\mathbf{Ad}   ( \boldsymbol{\mathcal{P}} (\alpha(\sigma(\mathbf{k}^c)))  )    \boldsymbol{\xi}_{\mathbf{k}^c}\right) 
	\mathbf{Exp}\left( \boldsymbol{\beta} \right)
	\\ & \approx 
	\mathbf{Exp} \left( 
	\sigma(\mathbf{k}^c) \mathbf{J}_{\mathbf{l}}^{-1} ( \boldsymbol{\beta} ) 
	\mathbf{Ad}  ( \boldsymbol{\mathcal{P}} (\alpha(\sigma(\mathbf{k}^c))) )  \boldsymbol{\xi}_{\mathbf{k}^c}
	+
	\boldsymbol{\beta}
	\right)
	.
	\end{aligned}
	\end{equation*}
	At last, the partial derivative could be calculated as follow,
	\begin{equation*}
	\begin{aligned}
	& \frac{\partial  \mathbf{Log}( {\boldsymbol{\mathcal{C}}}^{\mathrm{lhs}} ) }{\partial \boldsymbol{\xi}_{\mathbf{k}^c}^{T} }
	=
	\frac{\partial  \mathbf{Log}( {\boldsymbol{\mathcal{C}}}^{\mathrm{lhs}} \vert_{\mathbf{k}^c} ) }{\partial \boldsymbol{\xi}_{\mathbf{k}^c}^{T} }
	\\  = &
	\sigma(\mathbf{k}^c) \mathbf{J}_{\mathbf{l}}^{-1} ( \boldsymbol{\beta} ) 
	\mathbf{Ad}\left( \boldsymbol{\mathcal{P}} (\alpha(\sigma(\mathbf{k}^c))) \right)
	.
	\end{aligned}
	\end{equation*}

	It can be seen that $\mathbf{J}_{\mathbf{l}}^{-1} ( \boldsymbol{\beta} ) $ occurs for each derivative
	$
	{\partial  \mathbf{Log}( {\boldsymbol{\mathcal{C}}}^{\mathrm{lhs}} \vert_{\mathbf{k}^c} ) }/{\partial \boldsymbol{\xi}_{\mathbf{k}^c}}
	$,
	so the final linearized cycle writes
	\begin{equation}
	\label{eq. cycle-pgo linearied gemometric constraints.}
		\sum_{ \mathbf{k}^c = \mathbf{1}^c }^{\boldsymbol{\lambda}^c} 
		\sigma(\mathbf{k}^c)
		\mathbf{Ad}\left( \boldsymbol{\mathcal{P}} (\alpha(\sigma(\mathbf{k}^c))) \right)
		\boldsymbol{\xi}_{\mathbf{k}^c}
		+ \mathbf{J}_{\mathbf{l}} ( \boldsymbol{\beta} )  \boldsymbol{\beta}
		= 
		\mathbf{0}
		.
	\end{equation}
	
	Without loss of generality, let us assume ${\boldsymbol{\mathcal{C}}}^{\mathrm{lhs}} = \mathbf{I}$ is the $i$-th geometric cycle.
	Then the $i$-th block row and $\mathbf{k}^c$-th block column of $\mathbf{B}$ writes $\mathbf{B}_{i, \mathbf{k}^c}
	= \sigma(\mathbf{k}^c)
	\mathbf{Ad}\left( \boldsymbol{\mathcal{P}} (\alpha(\sigma(\mathbf{k}^c))) \right)$.
	The $i$-th block row of $\mathbf{b}$ writes $\mathbf{b}_i =   \mathbf{J}_{\mathbf{l}} ( \boldsymbol{\beta} )  \boldsymbol{\beta}$.

	For a set of cycles
	in a cycle basis,
	we linearize each one of them and assemble the results
	in the form of
	$
	\mathbf{B} \boldsymbol{\xi} + \mathbf{b} = \mathbf{0}
	$.

\section{Theorems and Proofs on Graph Theory}
\label{Appendix Proof of Theorem on graph theory}

	This section contains proofs of several key conclusions that have been used in the proposed MCB.
	Section \ref{proof of Theorem on the property of connected component} is the proof for Theorem \ref{thoerem: property of connected component}
	that is used for the construction of an isometric set.
	Theorem \ref{Theorem: rank relation restricted and full cycle incidence}
	is used for the restriction of cycle/support vectors.
	Lemma \ref{lemma: independence of cycles based on support vectors}
	is used in independence tests.

\subsection{Proof of Theorem \ref{thoerem: property of connected component}}
\label{proof of Theorem on the property of connected component}
	\begin{IEEEproof}
		It has been proved in \cite{amaldi2009breaking} that all representations of an isometric cycle $\mathcal{C}$ exactly correspond to a single connected component in $\mathcal{G}^{\dagger}$.
		We extend this result in what follows.

There are three possible switches in Lemma \ref{lemma isometric circuit identification},
i.e., case (1), case (2).(a) and case (2).(b).
		First, we observe that the switch in each case is mutual: If $\mathcal{C}$ can be switched to $\mathcal{C}'$, then $\mathcal{C}'$ can be switched to $\mathcal{C}$ by the same principle.
	For case (1), it is straightforward by the fact that $\mathcal{P}_{uv} = \mathcal{P}_{vu}$, so $\mathcal{C}(v, e_{uv}) = \mathcal{C}(u, e_{uv})$.
		In case (2).(a), we need to prove $\mathcal{C}(x, e_{uv}) = \mathcal{C}(x', e_{uv})$. Starting from $\mathcal{C}(x', e_{uv})$, $x$ is the first vertex on the shortest path from $x'$ to $v$, i.e. $x = s_{x'}(v)$. It can be verified that $x' = s_x(u)$, so according to Lemma \ref{lemma isometric circuit identification}.2.(a), $\mathcal{C}(x', e_{uv}) = \mathcal{C}(x, e_{uv})$.
		In Lemma \ref{lemma isometric circuit identification}.(2).(b), we prove $\mathcal{C}(x, e_{uv}) = \mathcal{C}(v, e_{xx'})$. Starting from $\mathcal{C}(v, e_{xx'})$, $u$ is the first vertex on the shortest path from $v$ to $x'$, i.e. $u = s_v(x')$. It can be verified that $v \neq s_u(x)$ and $x' = s_x(u)$, so by Lemma \ref{lemma isometric circuit identification}.(2).(b), $\mathcal{C}(v, e_{xx'}) = \mathcal{C}(x, e_{uv})$.
		This implies that in $\mathcal{G}^{\dagger}$, the arcs (i.e., directed edges) are mutual:
		If there is an arc from $\mathcal{C}$ to $\mathcal{C}'$, then there is an arc from $\mathcal{C}'$ to $\mathcal{C}$ as well.

		Second, for an isometric circuit $\mathcal{C}(x, e_{st})$, we can generate exactly two switches
		because in Lemma \ref{lemma isometric circuit identification}, the vertex $u$ can be chosen as either $s$ or $t$, and $v$ as either $t$ or $s$ accordingly.
		This implies that if a vertex in $\mathcal{G}^{\dagger}$ is a representation of an isometric circuit, it must have an out-degree $d_{out} = 2$.

	Now assume we replace the two arcs mutually connecting $\mathcal{C}$ and $\mathcal{C}'$ in $\mathcal{G}^{\dagger}$ by an undirected edge.
	Considering the fact that all representations of an isometric circuit lie in the same connected component of $\mathcal{G}^{\dagger}$ \cite{amaldi2009breaking},
	we conclude: By replacing mutual arc pairs in $\mathcal{G}^{\dagger}$ as undirected edges, all representations of an isometric circuit constitute a connected subgraph whose vertices have degree of $2$, which is a simple cycle.
	Therefore the directed version is a double-linked directed cycle in $\mathcal{G}^{\dagger}$.

		An isometric circuit has $\vert \mathcal{C} \vert$ representations, so the directed cycle has $\vert \mathcal{C} \vert$ vertices.
	\end{IEEEproof}

	\begin{theorem}
		\label{Theorem: rank relation restricted and full cycle incidence}
		Let $\mathcal{T}$ be any tree (not necessarily a spanning tree) in $\mathcal{G}(\mathcal{V}, \mathcal{E})$.
		Define $\bar{\mathcal{C}} \in \mathbb{Z}_2^{\nu}$ to be the incidence vector of $\mathcal{C}$ restricted to the set of off-tree edges in $\mathcal{E} \backslash \mathcal{T}$,
		i.e.,
		\begin{equation*}
		\mathcal{C} = [  \bar{\mathcal{C}}, \tilde{\mathcal{C}} ],
		\ \mathrm{where}
		\ \bar{\mathcal{C}} \in \mathcal{E} \backslash \mathcal{T},
		\ \tilde{\mathcal{C}} \in \mathcal{T}
		.
		\end{equation*}
		Then for a collection of cycles $\{  \mathcal{C}_i = [  \bar{\mathcal{C}}_i, \tilde{\mathcal{C}}_i ] \}_{i \in N}$,
		we have $\mathbf{rank}( \{ \mathcal{C}_i \}_{i \in N} )  = \mathbf{rank} (  \{ \bar{\mathcal{C}}_i \}_{i \in N} ) $.
	\end{theorem}
	\begin{IEEEproof}
		Note that $ \mathbf{rank} (  \{ \bar{\mathcal{C}}_i \}_{i \in N} ) \le  \mathbf{rank}( \{ \mathcal{C}_i \}_{i \in N} ) $.
		Then we verify the inequality shall never be reached. Otherwise, there exist at least a sequence of $\lambda_i$
		$(i \in K )$,
		such that
		$ \sum_{i \in K } \lambda_i \bar{\mathcal{C}}_i = \mathbf{0} $
		and
		$ \sum_{i \in K } \lambda_i \tilde{\mathcal{C}}_i \neq \mathbf{0} $.
		As a result, $\mathcal{C}' = \sum_{i \in K } \lambda_i \mathcal{C}_i $ becomes a tree, containing edges only in $\mathcal{T}$,
		which is impossible.
		(By composing cycles, the change of the degree for a vertex is even,
		which implies that each vertex in $\mathcal{C}'$ has an even degree, thus $\mathcal{C}'$ cannot be a tree.) 
	\end{IEEEproof}

	\begin{lemma}
		\label{lemma: independence of cycles based on support vectors}
		Let $\{ \bar{\mathcal{S}}_i \}_{j=k}^{\nu}$ be the support vectors of a collection of independent circuits $\{ \mathcal{C}_i \}_{i=1}^{k-1}$.
		Then a circuit $\mathcal{C}_k$ is linearly independent from $\{ \mathcal{C}_i \}_{i=1}^{k-1}$,
		if and only if there exists a $\bar{\mathcal{S}}_l$ $(k \le l \le \nu)$ such that $\langle \bar{\mathcal{C}}_k, \bar{\mathcal{S}}_l \rangle = 1$.
	\end{lemma}

	\begin{IEEEproof}
		\textsf{Sufficiency:}
		This is well-known in \cite{de1995applications, kavitha2008tilde, kavitha2009cycle}.
		Assume $\mathcal{C}_k$ is dependent. Then there is a non-trivial linear combination $\bar{\mathcal{C}}_k  = \sum_{i=1}^{k-1} \lambda_i \bar{\mathcal{C}}_i$.
		As a result, $\langle \bar{\mathcal{C}}_k, \bar{\mathcal{S}}_j \rangle = \sum_{i=1}^{k-1} \lambda_i \langle \bar{\mathcal{C}}_i, \bar{\mathcal{S}}_j \rangle = 0$ holds for all $k \le j \le \nu$,
		which contradicts the existence of $\bar{\mathcal{S}}_l$.
		\textsf{Necessity:}
		Assume $\langle \bar{\mathcal{C}}_k, \bar{\mathcal{S}}_j \rangle = 0$ for all $k \le j \le \nu$.
		Then the orthogonality $\langle \bar{\mathcal{C}}_i, \bar{\mathcal{S}}_j \rangle = 0$ would hold
		for all $1 \le i \le k$, $k \le j \le \nu$.		
		Since $\mathcal{C}_k$ is independent from  $\{ \mathcal{C}_i \}_{i=1}^{k-1}$, then $\{ \mathcal{C}_i \}_{i=1}^{k}$ are a set of independent circuits.
		By theorem \ref{Theorem: rank relation restricted and full cycle incidence},
		the restricted circuits $\{ \bar{\mathcal{C}}_i \}_{i=1}^{k}$ are also independent.
		Then $\{ \bar{\mathcal{C}}_i \}_{i=1}^{k} \cup \{ \bar{\mathcal{S}}_j \}_{j=k}^{\nu}$ are $\nu + 1$ linearly independent vectors for a space of dimension $\nu$,
		which cannot be true.
	\end{IEEEproof}

\section{Lexicographic Dijkstra (LexDijkstra)}
\label{appendix. consistent APSP. lexicographic djikstra}

	In this section, we provide the proofs related to LexDijkstra.
	To begin with, lexicographic paths are defined in Definition \ref{appendix def lexicographic ordering by edge ids}.
	Lemma \ref{appendix theorem: lexicographic paths to consistent APSP} builds the connection between lexicographic paths and consistent paths.
	Theorem \ref{appendix theorem Dijkstra property - all paths serached} is a supportive conclusion to derive LexDijkstra.
	The correctness of LexDijkstra is proved in Section \ref{proof of proposition: LexDijkstra}.
	Last but not least,
	Section \ref{proof of theorem on traverse back to a common node} is the proof for a key result that can greatly improve the effectiveness of lexicographic comparisons.


	\begin{definition}
		\label{appendix def lexicographic ordering by edge ids}
		(Lexicographic Paths \cite{hartvigsen1994all}).
		Consider undirected graph $\mathcal{G}(\mathcal{V}, \mathcal{E})$ with edge weight vector $w$. Each edge in $\mathcal{E}$ is assigned with a unique index, and a path $\mathcal{P}$ is described as a collection of edges.
		Then for every pair of nodes $u, v \in \mathcal{V}$, there exists a unique $u\ \textnormal{-}\ v$ path $\mathcal{P}_{uv}$, that satisfies exactly one of the following three conditions with respect to any other $u\ \textnormal{-}\ v$ path $\mathcal{P}'_{uv}$:
		\begin{itemize}
			\item[(1)] $w(\mathcal{P}_{uv}) < w(\mathcal{P}'_{uv})$,
			\item[(2)] $w(\mathcal{P}_{uv}) = w(\mathcal{P}'_{uv})$ and $\vert \mathcal{P}_{uv} \vert < \vert \mathcal{P}'_{uv} \vert$,
			\item[(3)] $w(\mathcal{P}_{uv}) = w(\mathcal{P}'_{uv})$, $\vert \mathcal{P}_{uv} \vert = \vert \mathcal{P}'_{uv} \vert$,
			and
			
			$\mathrm{min\_index} ( \mathcal{P}_{uv} \backslash \mathcal{P}'_{uv} ) <  \mathrm{min\_index} ( \mathcal{P}'_{uv} \backslash \mathcal{P}_{uv} ) $.
		\end{itemize}
	\end{definition}

	\begin{lemma}(\cite{hartvigsen1994all})
		\label{appendix theorem: lexicographic paths to consistent APSP}
		If all paths in a APSP are lexicographic paths, then the APSP is a consistent APSP.
	\end{lemma}

	\begin{IEEEproof}
		The proof is not given in \cite{hartvigsen1994all}, thus we provide one for completeness.
		Let $s$ and $t$ be two vertices on the lexicographic shortest path $\mathcal{P}_{uv}$.
		Suppose the lexicographic shortest path from $s$ to $t$, denoted by $\mathcal{P}_{st}$, is not on $\mathcal{P}_{uv}$.
		Let $\mathcal{P}_{st}^{\ast}$ be the subgraph on $\mathcal{P}_{uv}$ joining $s$ and $t$,
		i.e., $\mathcal{P}_{uv} = \mathcal{P}_{us} + \mathcal{P}_{st}^{\ast} + \mathcal{P}_{tv}$.
		Let us further define a path $\mathcal{P}'_{uv} = \mathcal{P}_{us} + \mathcal{P}_{st} + \mathcal{P}_{tv}$.
		Obviously, $w( \mathcal{P}_{st}^{\ast} ) = w (\mathcal{P}_{st})$ and $\vert \mathcal{P}_{st}^{\ast} \vert = \vert \mathcal{P}_{st} \vert$ (which implies $w(\mathcal{P}_{uv}) = w(\mathcal{P}'_{uv})$ and $\vert \mathcal{P}_{uv} \vert = \vert \mathcal{P}'_{uv} \vert$), otherwise either $\mathcal{P}_{uv}$ or $\mathcal{P}_{st}$ is not a lexicographic shortest path by the case (1) or case (2) of Definition \ref{appendix def lexicographic ordering by edge ids}.

		For the case (3) of Definition \ref{appendix def lexicographic ordering by edge ids}, we observe that,
		\begin{equation}
		\mathrm{min\_index}( \mathcal{P}_{uv} \backslash \mathcal{P}'_{uv} ) = \mathrm{min\_index}( \mathcal{P}_{st}^{\ast} \backslash \mathcal{P}_{st} )
		\label{eq. min_Index, path and subpath, condition 1}
		\end{equation}
		\begin{equation}
		\mathrm{min\_index}( \mathcal{P}'_{uv} \backslash \mathcal{P}_{uv} ) = \mathrm{min\_index}( \mathcal{P}_{st} \backslash \mathcal{P}_{st}^{\ast} ).
		\label{eq. min_Index, path and subpath, condition 2}
		\end{equation}
		Since $\mathcal{P}_{st}$ is the lexicographic shortest path, we have
		\begin{equation}
		\mathrm{min\_index}( \mathcal{P}_{st} \backslash \mathcal{P}_{st}^{\ast} )
		<
		\mathrm{min\_index}( \mathcal{P}_{st}^{\ast} \backslash \mathcal{P}_{st} )
		.
		\label{eq. min_Index, path and subpath, condition 3}
		\end{equation}
		By Eq. (\ref{eq. min_Index, path and subpath, condition 1})(\ref{eq. min_Index, path and subpath, condition 2})(\ref{eq. min_Index, path and subpath, condition 3}),
		we have $\mathrm{min\_index} ( \mathcal{P}'_{uv} \backslash \mathcal{P}_{uv} ) < \mathrm{min\_index} ( \mathcal{P}_{uv} \backslash \mathcal{P}'_{uv} ) $.
		Therefore by Definition \ref{appendix def lexicographic ordering by edge ids}.(3),
		$\mathcal{P}_{uv}$ is not a lexicographic shortest path since $\mathcal{P}'_{uv}$ is lexicographically shorter than $\mathcal{P}_{uv}$, which contradicts the assumption.
	\end{IEEEproof}

	\begin{theorem}
		\label{appendix theorem Dijkstra property - all paths serached}
		Let $\mathcal{G}$ be an undirected graph with weight $w(e) > 0, \forall e \in \mathcal{E}$.
		A single-source shortest path tree grown at $v$ can be computed using Dijkstra's algorithm \cite{dijkstra1959note}.
		Let $\mathcal{P}_{uv}$ be a shortest path from $u$ to $v$.
		Then for Dijkstra's algorithm,
		if $w(\mathcal{P}_{uv})$ comes to be the minimum weight available amongst that of all vertices whose shortest path has not been decided yet (such that $u$ is the next vertex to be expanded),
		all possible paths from $u$ to $v$ with weight $w(\mathcal{P}_{uv})$ have been traversed. 
	\end{theorem}
	\begin{IEEEproof}
		Let an arbitrary shortest path from $u$ to $v$ with the minimum weight $w(\mathcal{P}'_{uv}) = w(\mathcal{P}_{uv})$ be described by a vertex-edge alternating sequence
		$\mathcal{P}'_{uv} = \{ u\ \textnormal{-}\ e_{us}\ \textnormal{-}\ s, \cdots v \}$.
		Then there exists a path $\mathcal{P}_{sv} = \mathcal{P}'_{uv} \backslash e_{us}$ from $s$ to $v$ with weight $w(\mathcal{P}_{sv}) < w(\mathcal{P}_{uv})$ because $w(e_{us}) > 0$.
		When $w(\mathcal{P}_{uv})$ comes as the minimum weight amongst that of all vertices whose shortest path has not been decided yet,
		$\mathcal{P}_{sv}$ must have been considered, and a shortest path from $s$ to $v$ must have been found with weight at most $w(\mathcal{P}_{sv})$.
		Then at the expansion stage of node $s$, the edge $e_{us}$ has been traversed, so has been the path $\mathcal{P}'_{uv} = \mathcal{P}_{sv} \cup e_{us}$.
	\end{IEEEproof}

\subsection{Proof of Proposition \ref{appendix. proposition Lexicographic Dijkstra}}
\label{proof of proposition: LexDijkstra}
	
	\begin{IEEEproof}
		By Theorem \ref{appendix theorem Dijkstra property - all paths serached},
		all shortest paths from $u$ to $v$ with exactly the same minimum weight, which is the case (2) and (3) in Definition \ref{appendix def lexicographic ordering by edge ids},
		will be traversed by Dijkstra's algorithm before the final path $\mathcal{P}_{uv}$ is decided.
		Thus it would be sufficient to perform a ``lexicographic comparison" in Dijkstra's update process according to Definition \ref{appendix def lexicographic ordering by edge ids} whenever a new path from $u$ to $v$ is found.
		By Lemma \ref{appendix theorem: lexicographic paths to consistent APSP},
		the APSP comprising lexicographic paths constructed by Definition \ref{appendix def lexicographic ordering by edge ids} is consistent.
		Therefore $\mathcal{P}_{uv}$ constructed by Definition \ref{appendix def lexicographic ordering by edge ids} is a consistent path in a consistent APSP.

	\end{IEEEproof}

\subsection{Proof of Theorem \ref{theorem: LexDijkstra, only need to traverse back to a common node.}}
\label{proof of theorem on traverse back to a common node}
	
	\begin{IEEEproof}
		Let $\mathcal{P}_{uv}$ and $\mathcal{P}'_{uv}$ be two paths from $u$ to $v$.
		Let $x$ be an arbitrary common vertex shared by $\mathcal{P}_{uv}$ and $\mathcal{P}'_{uv}$.
		Then $\mathcal{P}_{uv}$ and $\mathcal{P}'_{uv}$ can be divided by $x$ into two parts:
		$
		\mathcal{P}_{uv} = \mathcal{P}_{ux} + \mathcal{P}_{xv}
		$,
		and
		$
		\mathcal{P}'_{uv} = \mathcal{P}'_{ux} + \mathcal{P}'_{xv}
		$.
Without loss of generality, let us assume the shortest path tree of LexDijkstra rooted at $v$.
Then by the time comparing paths between $u$ and $v$,
the lexicographic shortest path between $x$ and $v$ must have been found, and is unique, i.e., $\mathcal{P}'_{xv} = \mathcal{P}_{xv}$.
As a result, in case (3) of Definition \ref{appendix def lexicographic ordering by edge ids}, we have:
		$ \mathcal{P}_{uv} \backslash \mathcal{P}'_{uv}  = \mathcal{P}_{ux} \backslash \mathcal{P}'_{ux}$, and
		$ \mathcal{P}'_{uv} \backslash \mathcal{P}_{uv}  = \mathcal{P}'_{ux} \backslash \mathcal{P}_{ux}$.
Therefore it suffices to compare the paths (i.e., indices of edges) between $u$ and $x$.
Since $x$ is chosen arbitrary, we compare to the nearest shared vertex.

	\end{IEEEproof}

\end{appendices}


\begin{IEEEbiography}[{\includegraphics[width=1in,height=1.25in,clip,keepaspectratio]{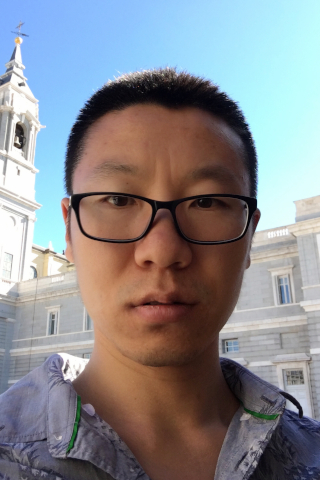}}]{Fang Bai}
	was born in Ningxia Province, China, in 1988.
	He received the computer science degree from Nankai University in 2010, and the Ph.D. degree in robotics from University of Technology Sydney in 2020.
	His Ph.D. thesis studies theoretical aspects in graph optimization, resulting a cycle based approach, and a closed form equation to predict the optimal values in least squares optimization.
	
	His research interests include both mathematical abstractions and practical applications in robotics.	
\end{IEEEbiography}
\begin{IEEEbiography}[{\includegraphics[width=1in,height=1.25in,clip,keepaspectratio]{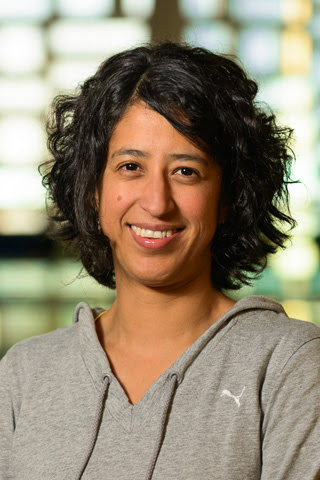}}]{Teresa Vidal-Calleja} received her BSc in Mechanical Engineering from the National Autonomous University of Mexico (UNAM), her MSc in Electrical Engineering from CINVESTAV-IPN, Mexico City, and her PhD in Automatic Control, Computer Vision and Robotics from the Polytechnic University of Catalonia (UPC), Spain in 2007. She was a postdoctoral research fellow at both, LAAS-CNRS in France and the Australian Centre for Field Robotics at the University of Sydney, Australia. She joined the Centre for Autonomous Systems at University of Technology Sydney (UTS) in 2012, where currently is Associate Professor. Her research interests are in robotic probabilistic perception.
\end{IEEEbiography}
\begin{IEEEbiography}[{\includegraphics[width=1in,height=1.25in,clip,keepaspectratio]{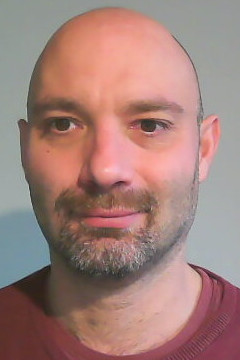}}]{Giorgio Grisetti}
is Associate Prof. at Sapienza University of Rome.
He achieved his Ph.D. from Sapienza University of Rome in 2006, and then conducted his post doctoral research
at the Autonomous Intelligent Systems Lab of University of Freiburg
from 2006 to 2010.
His research interests lie in SLAM, state estimation and navigation
for mobile robots.
He is author of more than 100 peer reviewed papers in journals and
conferences, and he is known for his contribution to open-source
packages addressing SLAM related problems such as GMapping, g2o, NICP,
Hog-Man, HBST and Pro-SLAM.
\end{IEEEbiography}

\end{document}